\documentclass{article} % For LaTeX2e
\usepackage{iclr2027_conference,times}

\usepackage{amsmath,amsfonts,bm,graphicx,subcaption,booktabs}

\def\eqref#1{equation~\ref{#1}}
\def\1{\bm{1}}

\DeclareMathAlphabet{\mathsfit}{\encodingdefault}{\sfdefault}{m}{sl}
\SetMathAlphabet{\mathsfit}{bold}{\encodingdefault}{\sfdefault}{bx}{n}

\usepackage{url}
\usepackage[colorlinks=true, allcolors=black]{hyperref}
\usepackage{xcolor}
\definecolor{citeblue}{RGB}{31,60,110}
\hypersetup{citecolor=citeblue, urlcolor=citeblue}
\usepackage{amsthm}
\usepackage{enumitem}
\usepackage{titletoc}
\usepackage{fontawesome5}

\theoremstyle{plain}

\iclrfinalcopy

\title{When Forgetting is not Catastrophic: \\On the Mechanics of Spurious Forgetting}

\author{%
  \hspace{-0.6em}
  Vedant Palit\textsuperscript{1,2,3} \,
  Florent Draye\textsuperscript{1,4} \,
  Nicolas Zucchet\textsuperscript{5} \,
  Zhijing Jin\textsuperscript{1,2,3} \,
  Bernhard Sch\"{o}lkopf\textsuperscript{1,6} \\[0.3em]
  \small\textsuperscript{1}MPI for Intelligent Systems, T\"{u}bingen \qquad
  \textsuperscript{2}Jinesis Lab, University of Toronto \& Vector Institute \\
    \small\textsuperscript{3}EuroSafeAI \qquad
  \textsuperscript{4}Hector Foundation \qquad
  \textsuperscript{5}Stanford University \qquad
  \textsuperscript{6}ELLIS Institute T\"{u}bingen \\[0.3em]
  \small\faGithub~\href{https://github.com/vedantpalit/spurious-forgetting-mechanics}{\texttt{vedantpalit/spurious-forgetting-mechanics}}}

\footnotetext{Corresponding author: \texttt{vedant.palit@tuebingen.mpg.de}}

\begin{document}

\maketitle

\begin{abstract}
Knowledge that a language model appears to forget during finetuning often remains stored and can be recovered, a phenomenon called spurious forgetting. Finetuning on new facts can even produce forgetting that undoes itself: recall of the old facts collapses, recovers as training continues on new facts alone, and only then erodes for good. We seek to understand when such forgetting is not catastrophic. A minimal associative memory reproduces these dynamics with three ingredients: keys with shared structure, concentrated new values, and normalization in the network. Finetuning moves all old representations along a common direction, hiding the old facts while preserving their relative geometry; normalization withdraws this shift once the new facts are learned, whereas fact-specific changes accumulate and cause the erosion. Moreover, subtracting the common shift eliminates the collapse in a Transformer trained on synthetic data, and removing a single direction from each weight update restores old facts in a pretrained language model. Forgetting thus combines a shared, reversible loss of access with a slow erosion of individual facts, and only the second is catastrophic. Which one dominates depends on whether the new data move old memories together or apart.

\end{abstract}
\section{Introduction}

Continually training a neural network on new data can severely damage the knowledge and skills it previously acquired, a phenomenon known as catastrophic forgetting \citep{MCCLOSKEY1989109,Ratcliff1990-RATCMO,FRENCH1999128}.
It has been reported across many settings, from image classification \citep{goodfellow2013empirical,zenke2017continual,ramasesh2021anatomy} to deep reinforcement learning \citep{kirkpatrick2017overcoming,rolnick2019experience}, and large language models are no exception \citep{luo2023empirical,zucchet2025how,marek2026forgetting}.
Yet, a loss of performance need not be a loss of knowledge: a different prompt can restore lost capabilities \citep{kotha2024understandingcatastrophicforgettinglanguage}, and brief training on a few old examples can undo forgetting that reflects a loss of task alignment rather than of knowledge, named spurious forgetting \citep{zheng2025spuriousforgettingcontinuallearning}.
Accuracy measured at a single point in training therefore cannot tell whether knowledge was erased or only hidden. What hides it, and whether training on the new data alone can reveal it again, remains unclear; we ask

\begin{center}
    \textit{When is forgetting not catastrophic?}
\end{center}

\begin{figure}[t]
  \centering
  \includegraphics[width=\linewidth]{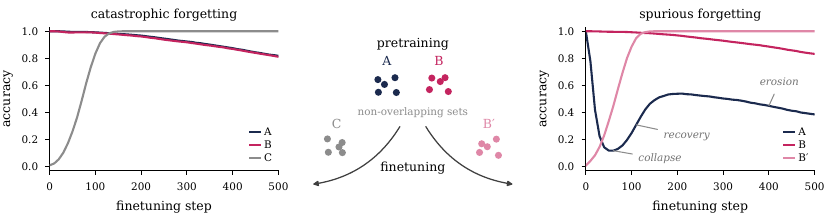}
  \vspace{-0.6cm}
\caption{\textbf{Changing only the finetuning data turns catastrophic forgetting into spurious forgetting.}
  A transformer is pretrained on two non-overlapping sets of facts, $A$ and $B$, and then finetuned on new facts only, without any replay.
  When the new facts overlap with both pretraining sets (left, finetuning on $C$), performance on both declines slowly and steadily. When they overlap only with $B$ (right, fine-tuning on $B'$), performance on $B$ is mostly intact, but performance on $A$ collapses, recovers although $A$ is never seen again, and later erodes.
  How the different sets are built and what it means for the new facts to overlap with the old ones is detailed in Section~\ref{sec:transformer}.}
  \label{fig:one}
\end{figure}

We study this question for factual knowledge, as forgetting and loss of access can be cleanly disentangled.
How language models store and retrieve facts is well understood \citep{geva2021transformer,meng2022locating,geva2023dissecting,nanda2023factfinding}, and synthetic biographies \citep{allenzhu2024physics31,zucchet2025how} let us specify exactly which facts the model knows, guarantee that the new facts were never seen before, and control how the two relate.
We can then ask not only whether the model still outputs an old fact, but also whether the fact is still stored.
In this setting, finetuning on new facts alone produces spurious forgetting in a particularly clear form (Figure~\ref{fig:one}): recall of the old facts collapses, then recovers as training continues on the same new facts, before a slower decline sets in.
Since there is no replay of the old facts, the recovery can only come from knowledge that survived the collapse.
\vspace{0.1cm}

Our contributions are as follows:
\vspace{-0.1cm}
\begin{itemize}[label=--, topsep=0pt, partopsep=0pt, parsep=0pt, itemsep=1pt, leftmargin=20pt]
    \item \textbf{A clean case of spurious forgetting.} In a controlled factual-recall setting, finetuning on new facts alone makes recall of the old facts collapse and then return, without any replay, and changing only where the new answers lie turns the same finetuning into steady forgetting.
    \item \textbf{A minimal mechanism.} A minimal associative memory reproduces the phenomenon once three ingredients are present: keys with shared structure, new answers concentrated in one region, and normalization.
    Finetuning then moves all old representations together, which hides the old facts without erasing them, and normalization undoes this shift once the new facts are learned.
    Changes specific to each fact accumulate and cause a slower, lasting forgetting.
    \item \textbf{Causal evidence in trained models.} In Transformers trained from scratch on synthetic data, removing the common shift eliminates the collapse, whose end is set by the learning of the new facts, not by the number of training steps. In a pretrained language model, recall of the old facts recovers only when the subjects of the new facts reveal nothing about their answers, and removing one direction of each weight update restores it beyond the trained model's recall at equal new-fact accuracy.
\end{itemize}
Together, our results show that forgetting is not catastrophic when the update moves old memories together rather than apart: the knowledge is then hidden rather than erased, and can be recovered, sometimes by just continuing finetuning for longer.

\section{Forgetting that undoes itself}
\label{sec:setup}

To understand when forgetting is not catastrophic, we first need a setting in which it clearly is not: one in which performance collapses, yet the knowledge demonstrably survives (Appendix ~\ref{app:rank}).
We use synthetic biographies \citep{allenzhu2024physics31}, on which spurious forgetting was identified \citep{zheng2025spuriousforgettingcontinuallearning}, as they let us control what the model knows before finetuning and what it learns during it. A transformer is pretrained on two non-overlapping sets of biographies, $A$ and $B$, in a single stage rather than one per set \citep{zheng2025spuriousforgettingcontinuallearning}, until it recalls all of their facts, and is then finetuned on biographies of new individuals, without any replay. We compare two sets of new biographies, $C$ and $B'$, which differ only in how they overlap with the old ones.
We describe how they are built, and what overlapping means precisely, in Section~\ref{sec:transformer}; for now, the only relevant property is that $C$ overlaps with both $A$ and $B$, while $B'$ overlaps only with $B$ (Figure~\ref{fig:one}, middle).

Finetuning on $C$ yields classical catastrophic forgetting behavior (Figure~\ref{fig:one}, left): as the new facts are learned, recall of both $A$ and $B$ declines slowly and steadily.
Finetuning on $B'$ barely affects recall of $B$, but recall of $A$ follows a strikingly different trajectory (Figure~\ref{fig:one}, right), in three phases that we refer to throughout the paper:
\vspace{0.1cm}
\begin{itemize}[label=--, topsep=1pt, partopsep=0pt, parsep=0pt, itemsep=1pt, leftmargin=20pt]
    \item \textbf{Collapse:} recall of $A$ quickly drops, well before the new facts are learned.
    \item \textbf{Recovery:} recall of $A$ then partially returns as the new facts are acquired, although $A$ never reappears in training.
    \item \textbf{Erosion:} finally, recall of $A$ declines slowly and persistently, as in catastrophic forgetting.
\end{itemize}

This trajectory raises two questions: why finetuning on $B'$ hides the facts of $A$ while sparing those of $B$, and why the collapse reverses on its own.
Section~\ref{sec:toy} answers both in a minimal model.
Section~\ref{sec:transformer} then returns to Transformers, and to the biography structure needed for the phenomenon to appear.

\section{A minimal model reveals the mechanism of spurious forgetting}
\label{sec:toy}

Factual recall, including that of synthetic biographies, can be abstracted as an associative memory: each fact maps a key, the subject it is about, to a value, its answer, which the model must retrieve when queried. We show that a minimal associative memory reproduces spurious forgetting once three ingredients are present: keys that share a common structure, new values concentrated in one part of the vocabulary, and a normalization layer. Its learning dynamics then explain the phenomenon: the structure of the data causes the collapse, and normalization drives the recovery. We check each step of this explanation in the model in Appendix~\ref{app:toy_checks}.

\subsection{Reproducing spurious forgetting in an associative memory model}

Many phenomena of factual recall, such as how many facts a model can store as a function of its size \citep{allenzhu2025physics33,nichani2025understanding}, how its loss scales \citep{cabannes2024scaling} or the dynamics with which facts are learned \citep{zucchet2025how, zucchet2026languagemodelssuffercurse}, can be reproduced in a simple feedforward associative memory, which strips away the sequence-modeling layers of the transformer.

\noindent\textbf{Associative memory task.}
Each fact $i$ is represented by a key
\begin{equation}
    k_i = \sqrt{\alpha}\,\mu + \sqrt{1-\alpha}\,e_i,
\end{equation}
where $\mu$ is a unit vector shared by all facts, $e_i$ a random unit vector specific to the fact, and $\alpha$ sets how much structure the keys share. Each key is associated with a value $y_i$, a token from a vocabulary of size $V$ that we split into two halves. The old facts form two sets, $A$, whose values are drawn uniformly from one half, and $B$, whose values are drawn from the other. The new facts $B'$ have new keys, sampled in the same way, and draw their values from the same half as $B$, just as the new biographies overlap with only one pretraining set in Figure~\ref{fig:one} (right); drawing them from the entire vocabulary instead mirrors finetuning on $C$ (left). The data thus combine keys that share structure with new values concentrated in one half of the vocabulary.

\noindent\textbf{Model.}
The model is a two-layer network with an RMS normalization layer \citep{zhang2019root} as the non-linearity between the two layers, followed by a softmax readout. It can be understood as the final linear layers of a Transformer. When queried with a key $k_i$, the model outputs a probability distribution over values,
\begin{equation}
    p_i=\mathrm{softmax}\!\left(W_2 \mathrm{RMS}(h_i) \right), \quad \text{with } h_i=W_1 k_i.
\end{equation}
The RMS normalization divides the hidden state by its norm and multiplies it by $\sqrt{d}$. The weight matrices $W_1$ and $W_2$ are trained with minibatch gradient descent on the cross-entropy loss, first on the pretraining data ($A$ and $B$) and then on the finetuning data ($B'$).

\begin{figure*}[t]
  \centering
  \includegraphics{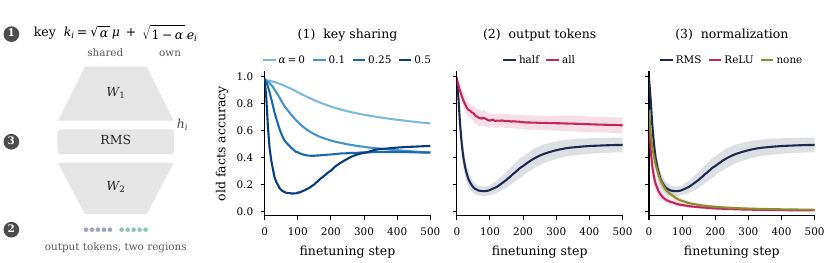}
  \vspace{-0.6cm}
\caption{\textbf{Shared keys, concentrated new values and normalization are all needed for the collapse, recovery and erosion.} The model maps keys, which share a component of strength $\alpha$, to values split into two regions, through one hidden layer with RMS normalization. As in Figure~\ref{fig:one}, it is pretrained on $A$ and $B$, whose values lie in different regions, then finetuned only on new facts $B'$, whose values lie in the region of $B$; curves show recall of $A$. Reducing $\alpha$ weakens the collapse and removes the recovery (1), drawing the new values from both regions leaves only gradual forgetting (2), and replacing the normalization removes the recovery (3). The model is described in Section~\ref{sec:toy}.}

  \label{fig:ingredients}
\end{figure*}

\noindent\textbf{Each of these ingredients is necessary.} We follow the old facts of $A$ over finetuning. The minimal model reproduces their collapse, recovery and erosion (Figure~\ref{fig:ingredients}), and removing any one of its ingredients breaks this trajectory.
With less shared structure in the keys, the collapse slows down until only gradual forgetting remains (panel 1). Drawing the values of the new facts from the entire vocabulary, as in finetuning on $C$, likewise leaves only gradual forgetting (panel 2). Without normalization, or with a ReLU in its place, the old facts collapse and never recover (panel 3).

\subsection{Key sharing leads to collapse}
\label{sec:suppression}

Intuitively, a gradient step on a new fact moves not only its own hidden state but that of every key resembling its own. All keys share a common component, so whatever the updates of the new facts have in common is written into all hidden states at once, the more so as $\alpha$ grows.
The result is a common shift that leaves the relative geometry of the hidden states mostly intact but, through the readout, favors the half of the vocabulary that contains the new answers, so that the facts of $A$ can no longer be decoded.

\noindent \textbf{All hidden states move in the same direction.}
When the model learns a new fact, gradient descent updates the first layer so that the hidden state of every key similar to the new fact's key moves against that fact's error.
Since all keys, old and new, share a common component, they are all pushed in the same direction whenever the errors of the new facts agree.
Formally, under gradient flow on the finetuning loss,
\begin{equation}
  \dot h_a = \dot{W}_1 k_a = -\frac{1}{n_B}\sum_{b} \big(k_b^\top k_a\big)\,\delta_b,
  \quad \text{where } \delta_b = \nabla_{h_b}\ell_b .
  \label{eq:write}
\end{equation}
In high dimension, the fact-specific parts of different keys are nearly orthogonal, so the overlap $k_b^\top k_a$ is close to its shared part $\alpha$ and, to leading order,
\begin{equation}
  \dot h_a = -\alpha\,\bar\delta_B,
  \quad \text{where } \bar\delta_B = \frac{1}{n_B}\sum_b \delta_b .
  \label{eq:meanwrite}
\end{equation}
The terms this approximation ignores vary in sign from one new fact to the next and largely cancel, so they are negligible compared with the one it keeps as long as the errors $\delta_b$ share a common component (Appendix~\ref{app:meanwrite}).
All the old hidden states then move together in the same direction (Figure~\ref{fig:mechanism}, left), at a speed proportional to $\alpha$: this is why the collapse slows down as $\alpha$ decreases and disappears at $\alpha = 0$ (Figure~\ref{fig:ingredients}, panel 1).

Whether the errors share a common component depends on the finetuning data: pretraining covers both halves of the vocabulary, so the model's predictions for the new keys are initially spread over both, whereas all the new answers lie in the same half, and on average the errors therefore all push probability toward this half.
When the new answers are instead drawn from the entire vocabulary, this common push largely cancels, and so does the collapse (Figure~\ref{fig:ingredients}, panel 2).

\noindent \textbf{The common shift hides the facts of $A$.}
The collapse of $A$ is a direct consequence of this common shift.
Ignoring normalization for a moment, shifting all hidden states by the same vector adds the same bias to the logits of every old fact, favoring the half of the vocabulary that contains the new answers.
The facts of $A$, whose answers lie in the other half, therefore lose to the new answers.
Normalization amplifies this effect.
Once the shift accounts for most of the norm of the hidden states, normalizing rotates all of them toward the shift direction: the old states become more similar to one another, the part of each state that encodes the fact's own answer is scaled down, and the predictions follow the shift toward the answers most common among the new facts. The facts of $B$, whose answers lie in the favored half, are largely spared by the shift itself, which raises the answers of their half roughly together, in proportion to how often each is a new answer: removing the shift from their hidden states barely changes their recall, whereas it restores that of $A$.
As their answers are also those of the new facts, they are instead exposed to the fact-specific learning of the new facts, which erodes them steadily, mostly after the collapse (Section~\ref{sec:recovery}). 

\subsection{Normalization drives recovery}
\label{sec:recovery}

Early in finetuning, all hidden states are pushed in the same direction, and this common shift causes the collapse of $A$.
Because the normalization fixes the norm of each hidden state, the growing shift comes at the expense of the individual parts, which encode each fact's own information and eventually become negligible.
At first, this also helps the new facts, whose individual parts point to wrong answers.
However, once the model has learned their individual associations, the shift drowns these as well and prevents the model from predicting the new facts correctly.
Finetuning then undoes the shift, both on the new facts and on the old facts, restoring the individual parts of every hidden state.
For the facts of $A$, these still carry most of the information that distinguishes them from one another, and the readout weights of their answers are barely changed by finetuning, as they only receive gradient while the model still puts probability on these answers for the new facts.
The facts of $A$ can therefore be read out again, and their recall partially recovers.

\noindent \textbf{Normalization reverses the shift.}
We track the common shift through the norm $s$ of the displacement of the old facts' mean hidden state $\bar h_A$ since the start of finetuning, and call its direction $\hat u_{\mathrm{sh}}$ the shared direction.
Through Equation~\ref{eq:meanwrite}, the dynamics of $s$ are governed by whether pushing the hidden states of the new facts along the shared direction improves their predictions or not.
Because of the normalization, such a push has two competing effects.
On the one hand, it adds shared component, which improves the predictions on the new facts in proportion to the shared margin $\bar\gamma_{\mathrm{sh}}$: the extra logit that the shared direction gives the correct answer compared with the model's current prediction, averaged over the new facts.
On the other hand, it scales their individual parts down, which changes their predictions in proportion to the individual margin $\bar\gamma_{\mathrm{ind}}$, the same quantity measured along their individual parts.
Assuming for simplicity that all new hidden states have the same coordinate $s$ along $\hat u_{\mathrm{sh}}$ and the same individual norm $\sigma$, we obtain, to leading order,
\begin{equation}
  \dot s = \alpha\,\sqrt{d}\, \frac{\sigma \left( \sigma \bar\gamma_{\mathrm{sh}} - s \bar\gamma_{\mathrm{ind}} \right)}{\left( s^2 + \sigma^2 \right)^{3/2}}, \quad \text{with } s(0) = 0.
  \label{eq:shift}
\end{equation}
The first term corresponds to the added shared component and the second to the smaller individual parts; their relative size determines whether the common shift grows or shrinks.
The derivation of this equation and the precise definitions of the two margins are given in Appendix~\ref{app:shift_full}.

\begin{figure}[t]
  \centering
  \includegraphics{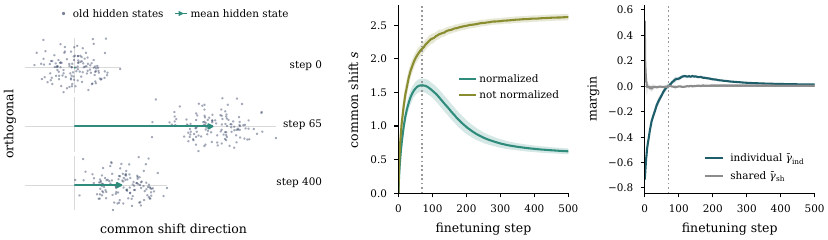}
  \vspace{-0.6cm}
  \caption{\textbf{Normalization reverses the shift responsible for forgetting.}
  The old representations first move together along a shared direction.
  Once the new facts are learned and the individual margin becomes positive, the common shift retracts and the old facts are recalled again.
  Without a normalization layer, this reversal does not occur, which is why the collapse is permanent (Figure~\ref{fig:ingredients}, panel 3).
  (left) Old hidden states before normalization, projected on the shared direction and an orthogonal one, after pretraining, at the end of the collapse and during erosion.
  (middle) Common shift magnitude $s$ of the old hidden states, with and without normalization.
  (right) Shared and individual margins of the new facts, which determine the dynamics of $s$ through Equation~\ref{eq:shift}.
  Same setup as in Figure~\ref{fig:ingredients}; the dotted line marks the end of the collapse.}
  \label{fig:mechanism}
\end{figure}

This equation explains the dynamics of the common shift in Figure~\ref{fig:mechanism}.
The shared margin only matters in the first steps of finetuning.
The shared direction can move probability between the two halves of the vocabulary, but not between the answers within a half: $\bar\gamma_{\mathrm{sh}}$ is positive at first, and once the predictions for the new facts have moved to the half of their answers, it can no longer improve them and remains close to zero (Figure~\ref{fig:mechanism}, right).
The shared margin thus sets the direction of the shift, and the individual margin then sets its magnitude.
The new facts are initially confidently wrong, $\bar\gamma_{\mathrm{ind}} < 0$, so scaling their individual parts down helps them: $s$ grows, and the growth reinforces itself.
Once the individual parts of the new facts point to their correct answers, $\bar\gamma_{\mathrm{ind}}$ becomes positive (Figure~\ref{fig:mechanism} right) and the same term pulls $s$ back (Figure~\ref{fig:mechanism} middle).
The shift therefore peaks when $\bar\gamma_{\mathrm{ind}}$ crosses zero.
This reversal waits for fact-specific learning, which is slow: the error of each new fact enters its own state with weight $1/n_B$, whereas their common component enters every state with weight $\alpha$ (Appendix~\ref{app:meanwrite}), in line with the general observation that shared structure is learned before item-specific structure \citep{Saxe_2019}. Without normalization, adding a shared component no longer scales the individual parts down, the $\bar\gamma_{\mathrm{ind}}$ term disappears in Equation~\ref{eq:shift}, and $\dot s = \alpha \bar\gamma_{\mathrm{sh}}$.
Nothing pulls the shift back (Figure~\ref{fig:mechanism} middle) and the collapse is permanent (Figure~\ref{fig:ingredients}).

\noindent \textbf{The facts of $A$ are hidden, not erased.}
Undoing the common shift only restores the facts of $A$ if what distinguishes them from one another has survived, and it does, for two reasons.
First, beyond the common shift, each old hidden state only receives the part of the update carried by the fact-specific parts of the new keys, which overlap weakly with its own key (Appendix~\ref{app:split}).
This individual displacement stays small during the collapse: while the mean of the old hidden states moves a lot, their relative geometry barely changes (Figure~\ref{fig:mechanism}, left). The facts of $A$ therefore keep their ordering within their own half, and even at the bottom of the collapse most of them still rank their correct answer first among the answers of that half (Appendix~\ref{app:toy_checks}).
Second, the readout weights of the answers of $A$ receive little gradient: they are only updated while the model puts probability on these answers for the new facts, which stops early in finetuning.
Removing the part of the update that acts on the shared direction, $\Delta W_1(t)\,\mu\mu^\top$, while keeping everything else, including the finetuned readout, restores their recall almost entirely (Figure~\ref{fig:app_shared_row}). 
The collapse is thus a loss of access to the facts of $A$, not a loss of the facts themselves.

The individual displacements nonetheless keep accumulating after the collapse, and they are not neutral: each old hidden state drifts toward the answers of the new facts whose keys overlap most with its own.
For $A$, restoring the mean hidden state therefore helps less and less as finetuning proceeds, and recovery remains incomplete.
For $B$, whose answers are those of the new facts, this drift, together with the readout weights of their half, which are refit to the new facts, erodes recall throughout finetuning, and removing the common shift barely helps.

\section{The mechanism in a controlled Transformer}
\label{sec:transformer}

Having identified the mechanism behind spurious forgetting in a minimal model, we now return to the Transformer of Figure~\ref{fig:one} and test whether its collapse and recovery follow the mechanism of the minimal model, first by showing that its data meet the same conditions, then through measurements and interventions on its representations. Table~\ref{tab:overview} lists where each analysis is repeated across models.

\subsection{Training Transformers on synthetic biographies meets the conditions}

\begin{figure*}[t]
  \centering
  \includegraphics{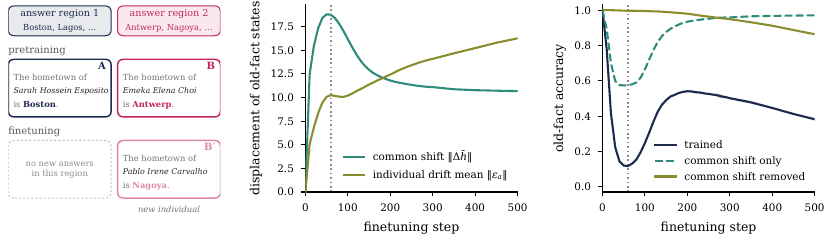}
  \vspace{-0.6cm}
  \caption{\textbf{The Transformer initially forgets through the same common shift as the minimal model.}
  (left) The Transformer is trained on synthetic biographies of fictitious individuals, each stating several facts about them, such as where they come from.
  The possible answers of each attribute are split into two subsets: the individuals of $A$ draw their answers from the first, and those of $B$ from the second.
  The model is pretrained on $A$ and $B$, then finetuned on new individuals $B'$ whose answers come from the same subset as $B$; this is the setup of Figure~\ref{fig:one} (right).
  (middle) The common shift $\Delta \bar h$ of the hidden states of $A$ grows fast and retracts at the end of the collapse, as in the minimal model (Figure~\ref{fig:mechanism}).
  The fact-specific displacements $\varepsilon_a$ grow more slowly and eventually become larger.
  (right) Accuracy on $A$, and when only the common part of the change of its logits is kept or removed.
  The dotted line indicates the end of the collapse.
  }
  \label{fig:transformer}
\end{figure*}

The minimal model identifies three conditions for spurious forgetting: a normalization layer, keys that share structure, and new answers concentrated in one region of the output space.
The first condition is met by any Transformer, as every residual block contains a normalization layer \citep{vaswani2017attention}.
In the pre-normalization architecture we use \citep{radford2019language,xiong2020layer}, the last one sits just before the readout.
The other two are properties of the data, which we now describe.
The synthetic biographies of Section~\ref{sec:setup} \citep{allenzhu2024physics31,zucchet2025how} describe a large population of fictitious individuals, each with a unique name and a few attributes, such as a hometown or an employer.
Each fact is stated by filling templates shared across all individuals, such as \texttt{The hometown of [name] is [city]} (Figure~\ref{fig:transformer}, left), and each set of biographies, $A$, $B$, $B'$ or $C$, contains its own individuals, so that the new facts are about individuals the model has never seen.
If we view the feedforward layers of a Transformer as an associative memory in which part of the context serves as the key and the next token as the value \citep{geva2021transformer}, the key of a fact is the name of the individual together with the attribute being queried, and its value is the answer.
As all facts about a given attribute are queried in a similar way, their keys share structure by design, just as the keys of the minimal model share the component $\mu$.
The last condition is not present by default, so we build it into the data: we split the possible answers of each attribute at random into two subsets.
The individuals of $A$ draw their answers from the first subset and those of $B$ from the second.
The finetuning sets therefore overlap with the old sets through their answers alone: those of $B'$ draw their answers from the second subset, like $B$, and those of $C$ from both subsets. Details of the data and the model are given in Appendix~\ref{app:setup}.

\subsection{The Transformer shows the same geometric split}

In the minimal model, the hidden states of the old facts first move together: their mean moves away from its pretrained value until the end of the collapse and then comes back, while the fact-specific displacements keep growing.
To test whether the same happens in the Transformer, we track the representations of the old facts of $A$ at the input of the readout, at the position where the model predicts the first token of the answer, for example \texttt{Boston} in Figure~\ref{fig:transformer} (left).
As in the minimal model, we decompose the displacement of each representation since the start of finetuning into a common shift $\Delta \bar h(t)$, its average over the facts of $A$ about the same attribute, and a fact-specific remainder $\varepsilon_a(t)$, whose size we measure by its average norm over the facts of $A$.

The two components evolve as the minimal model predicts (Figure~\ref{fig:transformer}, middle).
The common shift grows quickly, peaks around the bottom of the collapse, and then partly retracts during the recovery, as in Figure~\ref{fig:mechanism} (middle).
The fact-specific drift grows more slowly and, apart from a brief pause around the bottom of the collapse, keeps growing until it exceeds the common shift.

\subsection{Subtracting the common shift removes the collapse}

The minimal model predicts that the common shift alone causes the collapse: undoing it should restore the old facts, as long as what distinguishes them from one another has survived.
Since the readout is linear and shared by all facts, a common shift of the readout states adds the same change to the logits of all old facts. We therefore test this prediction on the logits, at the same position as above: we write the change of the logits of each old fact $a$ since the start of finetuning as $z_a(t)-z_a(0)=c(t)+v_a(t)$, where $c(t)$ is its average over the facts of $A$ about the same attribute, and measure recall from $z_a(0)+c(t)$, with only the common part, and from $z_a(t)-c(t)$, with the common part removed.

As predicted, subtracting the common part removes the collapse entirely (Figure~\ref{fig:transformer}, right): the old facts remain almost perfectly recalled, and only the slow decline caused by the fact-specific drift is left.
Keeping only the common part does the opposite and produces a collapse and a recovery by itself.
This collapse is shallower than in the trained model, because the common shift brings the old facts close to losing to the new answer region and the fact-specific drift pushes many of them across.
The collapse is therefore caused by the common shift, and erosion by the fact-specific drift.

The shift is carried mostly by the attention value and output weights, with little contribution from the query and key weights (Appendix~\ref{app:localization}).
In our attention-only architecture, these are the weights that write content into the residual stream, whereas the query and key weights only decide where attention goes, so this is consistent with the common shift being written into the associative memories of the network, as in the minimal model. With MLP blocks, the same three phases appear, and subtracting the common shift again removes the collapse (Appendix~\ref{app:mlp}).

\subsection{The learning rate changes the depth of the collapse}

In the minimal model, the common shift reverses once the individual margin of the new facts turns positive (Section~\ref{sec:recovery}), so the end of the collapse should be tied to the learning of the new facts rather than to a number of steps. Varying the learning rate and the batch size of finetuning confirms this (Appendix~\ref{app:dose}): the bottom of the collapse moves by a factor of three in steps, but falls at nearly the same accuracy on the new facts. This accuracy depends on how many new facts there are, however: with fewer, the bottom comes later in their learning, in the Transformer as in the minimal model, where the shift still peaks when the individual margin turns positive (Appendix~\ref{app:setsize}).

The depth of the collapse, in contrast, increases with the learning rate.
This effect lies outside our gradient-flow analysis, in which the learning rate only rescales time.
It is consistent with the observation that larger learning rates cause more forgetting, in continual pretraining \citep{gupta2023continual,ibrahim2024simple} as in finetuning at matched loss \citep{springer2025overtrained,lin2026sft,rofin2026learning}, and shows that part of this effect acts on forgetting that is spurious, and reverses once the new facts are learned.
\section{The collapse in a pretrained language model}
\label{sec:llm}

The controlled transformer meets the conditions of the minimal model by construction while a pretrained language model does not. We can still choose new facts whose answers the old facts never use, but the structure its keys share was set by pretraining on real text, so nothing guarantees that finetuning writes a common shift, or that it later withdraws it. The minimal model makes a prediction for each. The shift should be withdrawn only once the new facts have nothing left in common that helps predict their answers, so that each must be learned on its own; we test this with two kinds of new facts, one that meets this condition and one that does not. And since the shift of the hidden states is written by a single direction of the weight update, the one along which all new facts write together (Section~\ref{sec:recovery}), removing the top direction of each weight update should bring the old facts back.

\begin{figure*}[t]
  \centering
  \includegraphics[width=\linewidth]{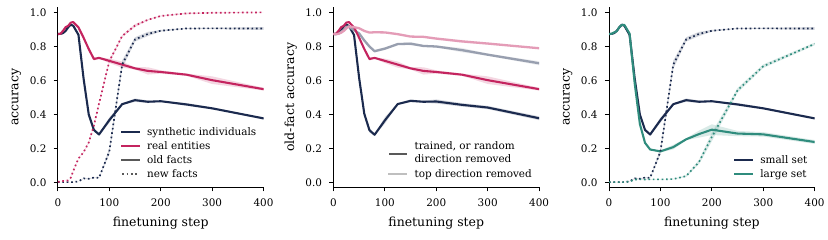}
\caption{\textbf{The collapse persists in a pretrained language model, but training reverses it only for arbitrary new facts.} OLMo~2~1B is finetuned without replay, in separate runs, on synthetic individuals or on real entities. Old facts (CounterFact facts the model knows) are solid, new facts dotted; three seeds. (left) Recall of the old facts drops for both kinds of new facts, but recovers only for synthetic individuals. (middle) Removing the top singular direction of each weight update (light) restores recall of the old facts for both kinds, whereas removing a random direction of the same norm leaves it where training did (dark). (right) With a larger set of synthetic individuals, the collapse goes deeper and lasts longer.}
\label{fig:llm}
\end{figure*}

\subsection{The collapse persists, and reverses when the new facts are arbitrary}

OLMo~2~1B \citep{olmo20252olmo2furious} knows far more facts than we can track, so we take as old facts the CounterFact facts \citep{meng2022locating} it answers correctly before finetuning; they play the role of $A$ (Appendix~\ref{app:llm}). We then finetune it without replay on documents stating new facts about people or products, in two separate runs with two kinds of new facts. In both, no new answer shares its first token with the answer of a tracked old fact, so the new answers occupy a region of the output space those old facts do not use, as the answers of $B'$ do for $A$. The two kinds differ in what their subjects say about their answers (Section~\ref{sec:recovery}). Synthetic individuals, invented names that receive real companies, cities, countries and languages at random, say nothing: as in the minimal model, the shared direction helps them reach the region of their answers, but not choose within it. Real entities, products from EntityQuestions \citep{sciavolino2021simple} whose manufacturer the model does not know, say something: the kind of product hints at the kind of manufacturer, as a car model hints at a carmaker, so the shared direction keeps helping within the region.

With both kinds of new facts, recall of the old facts first rises slightly, although none of them is replayed: finetuning on documents that state facts makes the model favor answers over other continuations, which fixes old facts it already ranked correctly among the possible answers (Appendix~\ref{app:llm_rise}). With synthetic individuals, the trajectory of the controlled settings then reappears (Figure~\ref{fig:llm}, left): recall of the old facts collapses before almost any new fact is recalled, recovers as the new facts are learned, and then erodes. More new facts delay the recovery, as each drives its own association with weight $1/n_B$ (Section~\ref{sec:recovery}): with a four times larger set of individuals, the collapse lasts longer and goes deeper (Figure~\ref{fig:llm}, right). With real entities, the old facts never recover and keep declining even after every new fact is recalled: the shared direction keeps helping the new facts choose their answers within the region, so the shift is never withdrawn.

\subsection{Removing one direction of the update restores the old facts}

In the controlled Transformer, we removed the common shift from the hidden states and logits of the old facts, which were all queried in the same way. In a pretrained model, the old facts are queried through varied prompts, at varied positions, so there is no single set of states over which to average, whereas the weights write the shift whatever the prompt. We therefore act on the weights. In the minimal model, the part of the update that all new facts share is rank-one, written along the shared key direction (Equation~\ref{eq:write}), and it initially dominates the update, while the errors of the new facts are aligned. In a deep network it need not be rank-one at any single layer, but it should still initially dominate each weight update, and thus concentrate in its top singular direction. In the minimal model and in both controlled Transformers, removing this direction indeed restores the old facts at the bottom of the collapse, and what it removes is aligned with the common shift (Appendices~\ref{app:toy_mirror} and~\ref{app:tf_lowrank}). We therefore subtract the top singular component of the update of every weight matrix in the Transformer blocks of the language model, or, as a control, a random rank-one matrix of the same norm, and evaluate the edited model (Appendix~\ref{app:lowrank}).

The edit largely removes the collapse (Figure~\ref{fig:llm}, middle), whereas the random control leaves recall exactly where training left it (Appendix~\ref{app:lowrank}). With synthetic individuals, only the slow erosion remains. With real entities, the edit restores most of the old facts that training never brought back: their steady decline was mostly a shift that was never withdrawn, not erasure. The removed direction also lowers recall of the new facts, but the edit does not rewind finetuning: once the new facts are partly learned, the edited model recalls more old facts than the trained model did at the same accuracy on the new facts (Appendix~\ref{app:lowrank}).

\section{Discussion}
\label{sec:discussion}

That a loss of performance need not be a loss of knowledge is an old idea: psychology separates stored from accessible information \citep{tulving1966availability,ebbinghaus1885gedachtnis}, and suppressed associations return once interference stops \citep{postman1968temporal}.
Connectionist networks show the same savings, and training them on unrelated associations can even restore forgotten ones \citep{hetherington1989there,hinton1987using,harvey1996unicycling,atkins1998recovery}.
In deep networks, linear probes often recover old tasks that the outputs no longer solve, a phenomenon called shallow forgetting \citep{murata2020happening,davari2022probing,hess2024knowledge,lanzillotta2025heads}; performance drops sharply at the start of a new task before returning with some replay \citep{lange2023continual,zucchet2025how}; and language models recover lost capabilities with a different prompt \citep{kotha2024understandingcatastrophicforgettinglanguage} or a few old examples, a case named spurious forgetting \citep{zheng2025spuriousforgettingcontinuallearning}.
In the settings we identify, the finetuning that hides the old facts also restores them, without probe, prompt or replay, through a shared shift that normalization withdraws once the new facts are learned. This shift towards the new answers may be one concrete form of the loss of task alignment that \citet{zheng2025spuriousforgettingcontinuallearning} identify behind spurious forgetting.

Forgetting is usually measured as a single quantity that grows with training, whereas our results suggest two processes on different timescales: a loss of access shared by all old facts, set by what the new data have in common, and a slower erosion specific to each fact.
The pretrained model also points to a trade-off between learning new facts quickly and recovering old ones. When the subjects of the new facts share something that predicts their answers, as the kind of a product predicts the kind of its manufacturer, the shared direction helps the model learn them, and they are learned faster. But for the same reason the shift along this direction keeps helping the new facts, so it is never withdrawn, and the old facts stay hidden. What helps a model generalize to new facts may thus be what keeps it from recovering the old ones.

\noindent\textbf{Limitations.}
Our minimal model is deliberately extreme, and normalization-driven recovery is likely weaker on real data, although the collapse may be more general. In the pretrained model, we attribute the difference in recovery to what the subjects reveal about their answers, but the two kinds of new facts also differ in other ways, and we study a single model.

% \subsection*{AI use statement}
% In this work, we used generative AI tools for to assist in the formalizing of proofs, with everything being reviewed by the authors. We have \textbf{not} used generative AI tools to generate synthetic data sets, develop theoretical models or conceptual frameworks, formulate mathematical claims, propose or refine hypotheses, provide feedback on research methodology or experiments, assist with translation, clean and reformat dataset, support qualitative and thematic data analysis, interpret results. Additionally, we used generative AI tools to edit code and modify scientific figures. We have reviewed all AI-assisted work. LLM-edited code was verified and tested for correctness by the authors. We take responsibility for the final content of this work, including text, claims or artifacts produced with the aid of generative AI.

\section*{Acknowledgment}
This work was supported in part by the German Federal Ministry of Education and Research (BMBF): Tübingen AI Center, FKZ: 01IS18039B; by the Deutsche Forschungsgemeinschaft (DFG, German Research Foundation) under Germany’s Excellence Strategy – EXC number 2064/1 – Project number 390727645; by Schmidt Sciences SAFE-AI Grant; by Coefficient Giving; and by the Hector Foundation. Resources used in preparing this research project were provided by the Max Planck Institute for Intelligent Systems and, in part, by the Province of Ontario, the Government of Canada through CIFAR, and companies sponsoring the Vector Institute.

% \subsection*{Reproducibility statement}

% Code to reproduce all experiments and figures is available in this \href{https://anonymous.4open.science/r/forgetting-dynamics-BEA7}{\textcolor{blue}{Repository}}, with instructions to run the minimal model, the transformer and the pretrained language model.
% Appendix~\ref{app:setup_all} describes the data, architectures and training of the transformer and of OLMo~2~1B, including how the old and new facts are built from publicly available datasets (CounterFact and EntityQuestions); Appendix~\ref{app:toy_notation} gives the sizes and training of the minimal model; and Appendices~\ref{app:transformer} and~\ref{app:llm_results} describe the procedure of every analysis and intervention.

\bibliography{iclr2027_conference}
\bibliographystyle{iclr2027_conference}

\appendix
\newpage
\startcontents[appendix]
\section*{Appendix}
{\hypersetup{linkcolor=citeblue}%
 \printcontents[appendix]{}{1}{\setcounter{tocdepth}{1}}}
\vspace{1em}
\noindent Table~\ref{tab:overview} lists where each analysis of the minimal model is repeated in the other models.
\begin{table}[h]
\centering\small
\begin{tabular}{lcccc}
\toprule
 & Minimal & \multicolumn{2}{c}{Transformer} & OLMo~2 \\
\cmidrule(lr){3-4}
Analysis & model & attention-only & with MLP blocks & 1B \\
\midrule
Collapse, recovery and erosion          & Fig.~\ref{fig:ingredients} & Fig.~\ref{fig:one} & App.~\ref{app:mlp} & Fig.~\ref{fig:llm} \\
New answers in both regions             & Fig.~\ref{fig:ingredients} & Fig.~\ref{fig:one} & App.~\ref{app:mlp} & -- \\
Common shift and fact-specific drift    & App.~\ref{app:toy_mirror} & Fig.~\ref{fig:transformer} & App.~\ref{app:mlp} & -- \\
Removing the common shift               & App.~\ref{app:shared_row} & Fig.~\ref{fig:transformer} & App.~\ref{app:mlp} & -- \\
Ordering within the region              & App.~\ref{app:toy_checks} & App.~\ref{app:rank} & App.~\ref{app:mlp} & -- \\
Weights that carry the shift            & -- & App.~\ref{app:localization} & App.~\ref{app:mlp} & -- \\
Learning rate and batch size            & App.~\ref{app:setsize} & App.~\ref{app:dose} & App.~\ref{app:mlp} & -- \\
Number of new facts                     & App.~\ref{app:setsize} & App.~\ref{app:setsize} & App.~\ref{app:setsize} & Fig.~\ref{fig:llm} \\
Removing the top direction of the update & App.~\ref{app:toy_mirror} & App.~\ref{app:tf_lowrank} & App.~\ref{app:tf_lowrank} & Fig.~\ref{fig:llm} \\
\bottomrule
\end{tabular}
\caption{Reference location for each analysis, for each model.}
\label{tab:overview}
\end{table}

\newpage
\section{Experimental setup}
\label{app:setup_all}

\begin{figure}[t]
  \centering
  \includegraphics{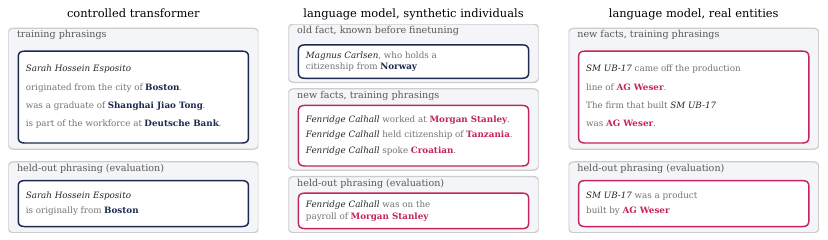}
  \caption{\textbf{Examples from the three datasets.} Each fact is learned from training phrasings and evaluated on held-out phrasings; answers in bold. The old facts of the language model are facts it already knows, and the new facts concern synthetic individuals, whose answers are unrelated to their names, or little-known real entities.}
  \label{fig:app_data}
\end{figure}

\subsection{Controlled transformer}
\label{app:setup}

\paragraph{Data.}
The transformer is trained on synthetic biographies \citep{allenzhu2024physics31,zucchet2025how}. Each individual has a unique name, made of a first, a middle and a last name, and six attributes: birth year, birthplace, university, field of study, employer and work city. Each attribute takes one value from a fixed pool of between 127 and 221 values. Every fact is stated in a sentence drawn from 25 phrasings of its attribute; for each individual and attribute, 20 phrasings are used for training and the remaining 5 are held out for evaluation. A biography consists of one sentence per attribute, in random order.

\paragraph{Old and new facts.}
The possible values of each attribute are split at random into two halves. Two pretraining populations of 2{,}000 individuals each draw their values from one half and from the other, respectively. Finetuning uses 500 new individuals, none of whom appear in pretraining, whose values are drawn either from the second half only (Figure~\ref{fig:one}, right) or from both halves (Figure~\ref{fig:one}, left).

\paragraph{Model and training.}
The model is an 8-layer transformer with model dimension 512, 8 attention heads, rotary position embeddings and pre-layer normalization, without MLP blocks and without dropout. It is pretrained for 16{,}000 steps with batches of 256
biographies, using AdamW ($\beta_1=0.9$, $\beta_2=0.95$, weight decay $0.1$, gradient clipping at $1$) with a cosine learning-rate schedule from $5\times10^{-4}$, until it recalls the pretraining facts perfectly. Finetuning
uses the same optimizer and batch size with a peak learning rate of $3.75\times10^{-5}$ over 1{,}200 steps, and sees only the new individuals. Results are averaged over three to five finetuning seeds.

\paragraph{Evaluation.}
Recall is the accuracy on the first token of the attribute value, predicted from the held-out phrasings of each fact. The within-region rank used in Appendix~\ref{app:rank} ranks the correct value among the values of its own half.

\subsection{Pretrained language model}
\label{app:llm}

\paragraph{Old facts.}
The old facts are CounterFact facts \citep{meng2022locating} that OLMo~2~1B \citep{olmo20252olmo2furious} answers correctly before finetuning, 1{,}106 in total across relations such as manufacturers, headquarters and citizenship.
Evaluation uses the 422 facts whose answer does not appear in the prompt, so that accuracy measures recall rather than copying.

\paragraph{New facts.}
The synthetic individuals have invented names and four attributes (employer, birth city, citizenship and language), with values drawn from pools of 66 real entities whose first tokens differ from those of the old answers. The large
finetuning set contains 1{,}000 individuals and the small set 250. The real entities are products from EntityQuestions \citep{sciavolino2021simple} with their manufacturers: 1{,}415 facts over 407 manufacturers, restricted to facts
the model does not know before finetuning, whose answers share no first token with the old answers and do not appear in the subject's name, with at most 60 facts per manufacturer. Each new fact is stated through 15 phrasings, 12 for training and 3 held out for evaluation.

\paragraph{Finetuning.}
Training documents are rendered afresh at every step from the training phrasings, with 32{,}768 tokens per step in sequences of 512 tokens. We use AdamW ($\beta_1=0.9$, $\beta_2=0.95$, no weight decay, gradient clipping at $1$) with a
learning rate of $10^{-5}$ after a 100-step linear warmup, and no replay of any pretraining data. Results are averaged over three seeds.

\paragraph{Evaluation.}
Recall is the top-1 accuracy of the first answer token over the full vocabulary, on the held-out phrasings for the new facts and on the CounterFact prompts for the old facts.

\clearpage

\newpage
\section{Analysis of the minimal model}
\label{app:toy}

This section derives the results of Sections~\ref{sec:suppression} and~\ref{sec:recovery}: the decomposition of the displacement of the old facts, how the new facts write into them, the dynamics of the common shift, and the size of the fact-specific displacements.
We also describe how we verify these results in simulations, and the interventions we run on the minimal model.

\subsection{Notation, assumptions and simulations}
\label{app:toy_notation}

Recall that each fact $i$ is represented by a key
\begin{equation}
    k_i = \sqrt{\alpha}\,\mu + \sqrt{1-\alpha}\,e_i \in \mathbb{R}^d,
\end{equation}
with $\mu$ a unit vector shared across all facts and $e_i$ drawn uniformly on the unit sphere, independently for each fact.
The model computes
\begin{equation}
    h_i = W_1 k_i, \quad x_i = \mathrm{RMS}(h_i) = \sqrt{d}\,\frac{h_i}{\lVert h_i \rVert}, \quad z_i = W_2 x_i, \quad p_i = \mathrm{softmax}(z_i),
\end{equation}
so that every normalized hidden state has norm $\sqrt{d}$.
During finetuning, the loss is the cross-entropy on the new facts only,
\begin{equation}
    L_B = \frac{1}{n_B} \sum_{b \in B} \ell_b, \quad \ell_b = -\log p_{b, y_b}.
\end{equation}
We use the index $a$ for old facts (set $A$) and $b$ for new facts (set $B'$, written $B$ in subscripts), bars for averages over a set (e.g. $\bar h_A = \frac{1}{n_A}\sum_a h_a$), hats for unit vectors (e.g. $\hat h = h / \lVert h \rVert$), and $\mathbf{1}_y$ for the one-hot vector of token $y$.
Time $t=0$ corresponds to the beginning of finetuning, and $\Delta W_1 = W_1(t) - W_1(0)$.

\paragraph{Assumptions.}
The analysis relies on four assumptions, each used only where stated:
\begin{enumerate}[label=Asm.\,\arabic*, left=0.4cm]
    \item \textbf{Gradient flow dynamics.} We study the gradient flow $\dot W_1 = -\nabla_{W_1} L_B$, $\dot W_2 = -\nabla_{W_2} L_B$ on the finetuning loss. Minibatch gradient descent with a small learning rate $\eta$, which we use in our simulations, approximates it, one step corresponding to a time increment $\eta$.
    \item \textbf{High-dimensional keys.} The overlap $u^\top v$ between two independent random unit vectors of $\mathbb{R}^d$ has zero mean and variance $1/d$. We therefore treat all the overlaps $e_i^\top e_j$ ($i \neq j$) and $\mu^\top e_i$ as small, of order $d^{-1/2}$, and only keep the leading-order terms in these overlaps.
    \item \textbf{The update does not depend on the old keys.} The old facts are not trained during finetuning, so $\Delta W_1$ only depends on their keys through the parameters at the end of pretraining. We neglect this dependence and treat the $e_a$ as independent of $\Delta W_1$. This assumption is only used in Section~\ref{app:laws}.
    \item \textbf{Common coordinate.} All new hidden states have the same coordinate along the shared direction $\hat u_{\mathrm{sh}}$ and the same norm orthogonal to it. This assumption is only used to obtain the closed-form Equation~\ref{eq:shift} (Section~\ref{app:shift_reduced}); the full dynamics do not rely on it.
\end{enumerate}
The readout $W_2$ does not appear in the old hidden states $h_a = W_1 k_a$.
All the results on the hidden states below therefore hold whether the readout is trained or not: $W_2$ only enters through the errors of the new facts, evaluated at its current value.
We return to the effect of the readout on the outputs of the old facts in Section~\ref{app:readout}.

\paragraph{Simulations.}
All simulations use $d = 128$, a vocabulary of $V = 32$ tokens split into two halves of 16, and $n = 128$ facts in each of $A$, $B$ and $B'$, with $\alpha = 0.5$ unless stated otherwise.
The model is pretrained on $A$ and $B$ with minibatches of 32 facts until both sets are recalled with accuracy at least $0.99$, then finetuned on $B'$ alone with minibatches of 32 new facts, at a learning rate chosen so that the new facts reach the same accuracy around step 200.
Results are averaged over ten seeds unless stated otherwise.

\subsection{Gradients of the finetuning loss}
\label{app:gradients}

We first compute the gradients of the finetuning loss with respect to the hidden states and to the two weight matrices, which the rest of the analysis uses throughout.

\paragraph{Error at the logits and at the normalized state.}
For the cross-entropy loss, the error of the new fact $b$ with respect to its logits is
\begin{equation}
    \nabla_{z_b} \ell_b = p_b - \mathbf{1}_{y_b} ,
\end{equation}
and, as $z_b = W_2 x_b$, its error with respect to the normalized state is
\begin{equation}
    \nabla_{x_b} \ell_b = W_2^\top \big(p_b - \mathbf{1}_{y_b}\big).
\end{equation}

\paragraph{Error at the hidden state.}
We denote by $\delta_b := \nabla_{h_b} \ell_b$ the error of the new fact at its hidden state.
Without normalization, $x_b = h_b$ and $\delta_b = W_2^\top \big(p_b - \mathbf{1}_{y_b}\big)$.
With normalization, the Jacobian of $h \mapsto \sqrt{d}\, h / \lVert h \rVert$ is
\begin{equation}
    \frac{\partial x}{\partial h} = \sqrt{d} \left( \frac{I}{\lVert h \rVert} - \frac{h h^\top}{\lVert h \rVert^3} \right) = \frac{\sqrt{d}}{\lVert h \rVert} \left( I - \hat h \hat h^\top \right).
\end{equation}
It is symmetric, so
\begin{equation}
    \delta_b = \frac{\sqrt{d}}{\lVert h_b \rVert} \left( I - \hat h_b \hat h_b^\top \right) W_2^\top \big(p_b - \mathbf{1}_{y_b}\big) ,
    \label{eq:app_delta}
\end{equation}
which is the expression of the main text.
The normalization has two effects on the error: it divides it by the norm of the state, and it removes its component along the state, $\hat h_b^\top \delta_b = 0$, since the loss does not depend on the norm of $h_b$.

The removed component has a simple interpretation.
As $\hat h_b^\top W_2^\top \big(p_b - \mathbf{1}_{y_b}\big) = d^{-1/2} x_b^\top W_2^\top (p_b - \mathbf{1}_{y_b}) = d^{-1/2} z_b^\top (p_b - \mathbf{1}_{y_b})$, we have
\begin{equation}
    \delta_b = \frac{\sqrt{d}}{\lVert h_b \rVert}\, W_2^\top \big(p_b - \mathbf{1}_{y_b}\big) + \frac{\gamma_b}{\lVert h_b \rVert}\, \hat h_b, \quad \text{with } \gamma_b := z_{b, y_b} - p_b^\top z_b .
    \label{eq:app_rms_split}
\end{equation}
Here $\gamma_b$ is the confidence margin of the new fact: the logit of the correct answer minus the average logit under the model's prediction.
It is negative when the model confidently predicts a wrong answer and positive when it predicts the correct one.
The first term is, up to a rescaling, the error a model without normalization would receive.
The second term points along the new fact's own state, with a sign given by its margin; for the new fact itself, it cancels the component of the first term along $\hat h_b$.

\paragraph{Gradients with respect to the weights.}
As $h_b = W_1 k_b$, the gradient of the loss of the new fact $b$ with respect to $W_1$ is the outer product of its error and its key, $\nabla_{W_1} \ell_b = \delta_b k_b^\top$, and averaging over the new facts gives
\begin{equation}
    \nabla_{W_1} L_B = \frac{1}{n_B} \sum_b \delta_b k_b^\top .
    \label{eq:app_gradW1}
\end{equation}
Similarly, as $z_b = W_2 x_b$,
\begin{equation}
    \nabla_{W_2} L_B = \frac{1}{n_B} \sum_b \big(p_b - \mathbf{1}_{y_b}\big)\, x_b^\top .
    \label{eq:app_gradW2}
\end{equation}
As $W_2$ sits after the normalization, its gradient does not involve the Jacobian above: unlike $W_1$, it receives no term along the new facts' states.

\subsection{Decomposition of the displacement of the old facts}
\label{app:split}

The keys are fixed and the old facts are not trained, so the displacement of an old hidden state is entirely due to the change of $W_1$:
\begin{equation}
    h_a(t) - h_a(0) = \Delta W_1 k_a = \underbrace{\sqrt{\alpha}\,\Delta W_1 \mu}_{\text{common shift}} + \underbrace{\sqrt{1-\alpha}\,\Delta W_1 e_a}_{\text{individual displacement}}.
    \label{eq:split}
\end{equation}
This decomposition is exact.
Averaging it over the old facts gives
\begin{equation}
    \bar h_A(t) - \bar h_A(0) = \sqrt{\alpha}\,\Delta W_1 \mu + \sqrt{1-\alpha}\,\Delta W_1 \bar e_A,
\end{equation}
and the deviation of each old fact from this mean motion is
\begin{equation}
    \varepsilon_a := \big(h_a(t) - h_a(0)\big) - \big(\bar h_A(t) - \bar h_A(0)\big) = \sqrt{1-\alpha}\,\Delta W_1 (e_a - \bar e_A).
    \label{eq:app_eps}
\end{equation}
The mean displacement thus equals the common shift $\sqrt{\alpha}\,\Delta W_1 \mu$ up to the term $\sqrt{1-\alpha}\,\Delta W_1 \bar e_A$.
As $\bar e_A$ is the average of $n_A$ independent random unit vectors, $\mathbb{E}\lVert \bar e_A \rVert^2 = 1/n_A$, and the calculation of Section~\ref{app:laws} shows that this term is a factor $\sqrt{n_A}$ smaller than a typical individual displacement $\varepsilon_a$.
We therefore measure the common motion through the norm of the mean displacement, $s = \lVert \bar h_A(t) - \bar h_A(0) \rVert$ (Figure~\ref{fig:mechanism}, middle), and the fact-specific motion through the spread
\begin{equation}
    \varepsilon_\mathrm{rms}(t) := \left( \frac{1}{n_A} \sum_{a} \lVert \varepsilon_a(t) \rVert^2 \right)^{1/2}
\end{equation}
(Figure~\ref{fig:app_erosion}).
When $\alpha = 0$, the common shift vanishes identically, and the whole displacement is fact-specific.

\subsection{How the new facts write into the old hidden states}
\label{app:meanwrite}

\paragraph{The write of a single new fact.}
Under the gradient flow dynamics, Equation~\ref{eq:app_gradW1} gives
\begin{equation}
    \dot W_1 = -\nabla_{W_1} L_B = -\frac{1}{n_B} \sum_b \delta_b k_b^\top ,
\end{equation}
so each new fact writes its error $\delta_b$ into $W_1$ along its own key.
The old facts are not part of the finetuning loss, but their hidden states move with $W_1$, and as their keys are fixed,
\begin{equation}
    \dot h_a = \frac{\mathrm{d}}{\mathrm{d}t}\big(W_1 k_a\big) = \dot W_1 k_a = -\frac{1}{n_B} \sum_b \big(k_b^\top k_a\big)\, \delta_b ,
    \label{eq:app_ha}
\end{equation}
where we used that $k_b^\top k_a$ is a scalar.
Each new fact thus adds its error $\delta_b$ to every old state, weighted by the overlap of their keys.
This equation is exact.

\paragraph{Key overlaps.}
Expanding the keys and using $\mu^\top \mu = 1$, the overlap between a new and an old key is
\begin{equation}
    k_b^\top k_a = \alpha + \sqrt{\alpha(1-\alpha)}\,\big(\mu^\top e_a + \mu^\top e_b\big) + (1-\alpha)\, e_b^\top e_a .
\end{equation}
Under Assumption~2, the last three terms have zero mean and are of order $d^{-1/2}$, so every pair of keys overlaps by approximately $\alpha$.
For the same reason, $\lVert k_b \rVert^2 = 1 + 2\sqrt{\alpha(1-\alpha)}\,\mu^\top e_b \simeq 1$.

\paragraph{The mean write.}
Equation~\ref{eq:app_ha} is linear in $k_a$, so averaging it over the old facts amounts to replacing $k_a$ by their mean key:
\begin{equation}
    \dot{\bar h}_A = \frac{1}{n_A} \sum_a \dot h_a = -\frac{1}{n_B} \sum_b \big(k_b^\top \bar k_A\big)\, \delta_b, \quad \text{with } \bar k_A = \sqrt{\alpha}\,\mu + \sqrt{1-\alpha}\,\bar e_A .
    \label{eq:app_mean_exact}
\end{equation}
The average leaves the shared vector $\mu$ untouched and reduces the fact-specific parts to $\bar e_A$.
The overlap of a new key with the mean old key is
\begin{align}
    k_b^\top \bar k_A &= \big(\sqrt{\alpha}\,\mu + \sqrt{1-\alpha}\,e_b\big)^\top \big(\sqrt{\alpha}\,\mu + \sqrt{1-\alpha}\,\bar e_A\big) \nonumber \\
    &= \alpha + \sqrt{\alpha(1-\alpha)}\,\mu^\top \bar e_A + \sqrt{\alpha(1-\alpha)}\,\mu^\top e_b + (1-\alpha)\, e_b^\top \bar e_A \nonumber \\
    &= \alpha\,(1+\xi) + \zeta_b ,
\end{align}
where we grouped the terms that are the same for all new facts into
\begin{equation}
    \xi := \sqrt{\frac{1-\alpha}{\alpha}}\,\mu^\top \bar e_A ,
\end{equation}
and the terms that depend on the new fact into
\begin{equation}
    \zeta_b := \sqrt{\alpha(1-\alpha)}\,\mu^\top e_b + (1-\alpha)\, e_b^\top \bar e_A .
\end{equation}
Plugging this decomposition into Equation~\ref{eq:app_mean_exact} and using $\frac{1}{n_B}\sum_b \delta_b = \bar\delta_B$ gives
\begin{equation}
    \dot{\bar h}_A = \underbrace{-\alpha\,(1+\xi)\,\bar\delta_B}_{\text{shared}} \; \underbrace{- \, \frac{1}{n_B} \sum_b \zeta_b\, \delta_b}_{\text{specific}} .
    \label{eq:app_mean_split}
\end{equation}
The sizes of the two terms are as follows.
\begin{itemize}
    \item[--] \textbf{Shared term.} All new facts contribute to it with the same coefficient $\alpha(1+\xi)$. As $\bar e_A$ is the average of $n_A$ independent random unit vectors, $\mu^\top \bar e_A$ has zero mean and variance $1/(n_A d)$, so $\xi$ is of order $(n_A d)^{-1/2}$ and the coefficient is $\alpha$ to leading order.
    \item[--] \textbf{Specific term.} $\mu^\top e_b$ has variance $1/d$ and $e_b^\top \bar e_A$ has variance $1/(n_A d)$, so the coefficients $\zeta_b$ have zero mean and are of order $d^{-1/2}$. They are moreover uncorrelated across new facts, as $e_b$ and $e_{b'}$ are independent for $b \neq b'$. Treating them, to leading order, as independent of the errors $\delta_b$ gives
    \begin{equation}
        \mathbb{E}\left\lVert \frac{1}{n_B} \sum_b \zeta_b\, \delta_b \right\rVert^2 = \frac{1}{n_B^2} \sum_b \mathbb{E}\big[\zeta_b^2\big]\, \lVert \delta_b \rVert^2 = O\!\left( \frac{\lVert \delta \rVert_\mathrm{rms}^2}{n_B\, d} \right),
    \end{equation}
    with $\lVert \delta \rVert_\mathrm{rms}^2 = \frac{1}{n_B}\sum_b \lVert \delta_b \rVert^2$. The cross terms vanish because the coefficients are uncorrelated: the $n_B$ contributions add up like a random walk, and their sum grows like $\sqrt{n_B}$ instead of $n_B$.
\end{itemize}
The specific term is therefore negligible compared with the shared one as soon as $\alpha \lVert \bar\delta_B \rVert \gg \lVert \delta \rVert_\mathrm{rms} / \sqrt{n_B d}$, that is, as soon as the errors of the new facts share a common component.
Keeping only the leading order of the shared term gives Equation~\ref{eq:meanwrite},
\begin{equation}
    \dot{\bar h}_A \simeq -\alpha\,\bar\delta_B ,
\end{equation}
and when $\alpha = 0$, only the specific term is left.

\paragraph{Why the errors of the new facts share a common component.}
The shared term is only large if the errors $\delta_b$ do not cancel when averaged over the new facts.
Averaging the decomposition of Equation~\ref{eq:app_rms_split} over the new facts gives
\begin{equation}
    \bar\delta_B = \frac{1}{n_B} \sum_b \frac{\sqrt{d}}{\lVert h_b \rVert}\, W_2^\top \big(p_b - \mathbf{1}_{y_b}\big) + \frac{1}{n_B} \sum_b \frac{\gamma_b}{\lVert h_b \rVert}\, \hat h_b .
\end{equation}
Early in finetuning, both terms survive the average, for two reasons that both come from the structure of the data:
\begin{itemize}
    \item[--] \textbf{The new answers all lie in the same region.} Up to the variations of the prefactor $\sqrt{d}/\lVert h_b \rVert$ across new facts, the first term is proportional to $W_2^\top (\bar p_B - q_B)$, where $\bar p_B$ is the average prediction on the new facts and $q_B = \frac{1}{n_B}\sum_b \mathbf{1}_{y_b}$ the distribution of their answers. All new answers belong to the same half of the vocabulary, so $q_B$ puts all its mass on that half. At the beginning of finetuning, the model has no reason to favor this half for the new keys: it was pretrained on answers from both halves, and its predictions for unseen keys spread over both. $\bar p_B - q_B$ therefore has a large component that moves probability mass from one half to the other, and this component is the same for all new facts. By contrast, which answer is correct within the half, and which wrong answer the model predicts, change from one new fact to the other and cancel in the average. When the new answers are drawn from the whole vocabulary, $q_B$ and $\bar p_B$ spread their mass over the two halves in a similar way and this component largely disappears, which is why the collapse does too (Figure~\ref{fig:ingredients}, panel 2).
    \item[--] \textbf{The new facts are confidently wrong.} At the beginning of finetuning, the model confidently predicts wrong answers for the new facts, so their margins $\gamma_b$ are all negative. At the same time, all new hidden states contain the shared component $\sqrt{\alpha}\, W_1 \mu$. The second term is thus a sum of vectors that share a direction, weighted by coefficients of the same sign, and it does not cancel either. This term gives rise to the $\gamma_{\mathrm{ind},b}$ contribution of Section~\ref{app:shift_full}.
\end{itemize}
Both contributions vanish as the new facts are learned: when $p_b \to \mathbf{1}_{y_b}$, $W_2^\top (p_b - \mathbf{1}_{y_b}) \to 0$ and $\gamma_b \to 0$.
The common component of the errors, and with it the shared write, therefore only exists early in finetuning.

\paragraph{Why the common write is faster than fact-specific learning.}
The same calculation applied to a new fact $b$ separates its own contribution from that of the other new facts:
\begin{equation}
    \dot h_b = -\frac{1}{n_B} \lVert k_b \rVert^2\, \delta_b - \frac{1}{n_B} \sum_{b' \neq b} \big(k_{b'}^\top k_b\big)\, \delta_{b'} \simeq -\frac{1}{n_B}\, \delta_b - \alpha\,\bar\delta_B ,
\end{equation}
where we used $\lVert k_b \rVert^2 \simeq 1$.
The error of a single fact enters its own state with weight $1/n_B$, whereas the common component of all errors enters every state with weight $\alpha$.
Whatever the errors of the new facts have in common is therefore learned much faster than what is specific to each, which is why the reversal of the common shift waits for fact-specific learning.

\subsection{Dynamics of the common shift}
\label{app:shift_full}

The normalization term of Equation~\ref{eq:app_rms_split} cancels along $\hat h_b$ for the new fact itself, but not for the old facts: through Equation~\ref{eq:app_ha}, it moves every old state along the parts that its state shares with $h_b$.
We make this precise by following the displacement of the old facts' mean state.
Let $s = \lVert \bar h_A(t) - \bar h_A(0) \rVert$ be its length, which we call the shared coordinate, and $\hat u_{\mathrm{sh}} = (\bar h_A(t) - \bar h_A(0)) / s$ its direction, which we call the shared direction.
As $\frac{\mathrm{d}}{\mathrm{d}t} \lVert v \rVert = \hat v^\top \dot v$ for any vector $v$, and $\bar h_A(0)$ is constant, Equation~\ref{eq:meanwrite} gives
\begin{equation}
    \dot s = \hat u_{\mathrm{sh}}^\top \dot{\bar h}_A \simeq -\frac{\alpha}{n_B} \sum_b \hat u_{\mathrm{sh}}^\top \delta_b .
\end{equation}

\paragraph{Validity of the approximation.}
Equation~\ref{eq:meanwrite} neglects the specific term of Equation~\ref{eq:app_mean_split}, and only its projection onto $\hat u_{\mathrm{sh}}$ matters for the dynamics of $s$.
The shared term contributes $\alpha$ times the mean of the projected errors $\hat u_{\mathrm{sh}}^\top \delta_b$, while the specific term, whose coefficients are random and of order $d^{-1/2}$, contributes a quantity of the order of their root mean square divided by $\sqrt{n_B d}$.
The approximation therefore requires the projected errors to share a sign across new facts, not the errors to be large.
This holds while the new facts are all wrong and once they are all learned, even when their errors become small and mostly fact-specific.
It only fails around the peak of $s$, where the projected errors change sign at different times for different new facts and $\dot s$ is small anyway.

\paragraph{Projecting the error.}
We decompose the direction of each new state into its component along $\hat u_{\mathrm{sh}}$ and an orthogonal component,
\begin{equation}
    \hat h_b = \cos\theta_b\, \hat u_{\mathrm{sh}} + \sin\theta_b\, \hat u_{\mathrm{ind},b} ,
\end{equation}
where $\theta_b$ is the angle between $h_b$ and $\hat u_{\mathrm{sh}}$, and $\hat u_{\mathrm{ind},b}$ is the unit vector along the individual part $h_b - (\hat u_{\mathrm{sh}}^\top h_b)\,\hat u_{\mathrm{sh}}$.
Then
\begin{equation}
    \begin{aligned}
    \hat u_{\mathrm{sh}}^\top \left( I - \hat h_b \hat h_b^\top \right) &= \hat u_{\mathrm{sh}}^\top - \cos\theta_b\, \hat h_b^\top \\
    &= \hat u_{\mathrm{sh}}^\top - \cos^2\theta_b\, \hat u_{\mathrm{sh}}^\top - \cos\theta_b \sin\theta_b\, \hat u_{\mathrm{ind},b}^\top \\
    &= \sin^2\theta_b\, \hat u_{\mathrm{sh}}^\top - \sin\theta_b\cos\theta_b\, \hat u_{\mathrm{ind},b}^\top ,
    \end{aligned}
\end{equation}
so that
\begin{equation}
    \hat u_{\mathrm{sh}}^\top \delta_b = \frac{\sqrt{d}}{\lVert h_b \rVert} \left( \sin^2\theta_b\, \hat u_{\mathrm{sh}}^\top W_2^\top \big(p_b - \mathbf{1}_{y_b}\big) - \sin\theta_b \cos\theta_b\, \hat u_{\mathrm{ind},b}^\top W_2^\top \big(p_b - \mathbf{1}_{y_b}\big) \right).
\end{equation}

\paragraph{Two margins.}
The two projections of the error are margins:
\begin{equation}
    \hat u_{\mathrm{sh}}^\top W_2^\top \big(p_b - \mathbf{1}_{y_b}\big) = (W_2 \hat u_{\mathrm{sh}})^\top \big(p_b - \mathbf{1}_{y_b}\big) = -\gamma_{\mathrm{sh},b}, \quad \text{with } \gamma_{\mathrm{sh},b} := (W_2 \hat u_{\mathrm{sh}})_{y_b} - p_b^\top W_2 \hat u_{\mathrm{sh}} ,
\end{equation}
and similarly $\hat u_{\mathrm{ind},b}^\top W_2^\top \big(p_b - \mathbf{1}_{y_b}\big) = -\gamma_{\mathrm{ind},b}$ with $\gamma_{\mathrm{ind},b} := (W_2 \hat u_{\mathrm{ind},b})_{y_b} - p_b^\top W_2 \hat u_{\mathrm{ind},b}$.
The shared margin $\gamma_{\mathrm{sh},b}$ is the margin by which the readout of the shared direction favors the correct answer of $b$ over the model's current prediction, and the individual margin $\gamma_{\mathrm{ind},b}$ is the same quantity for the individual direction of $b$.
As $z_b = \sqrt{d}\,( \cos\theta_b\, W_2 \hat u_{\mathrm{sh}} + \sin\theta_b\, W_2 \hat u_{\mathrm{ind},b} )$, the two margins decompose the confidence margin $\gamma_b$ of Equation~\ref{eq:app_rms_split} into a shared and an individual contribution,
\begin{equation}
    \gamma_b = \sqrt{d} \left( \cos\theta_b\, \gamma_{\mathrm{sh},b} + \sin\theta_b\, \gamma_{\mathrm{ind},b} \right).
\end{equation}
In particular, $\gamma_{\mathrm{ind},b}$ is negative while the individual part of $b$ points to a wrong answer the model is confident about, and becomes positive once it points to the correct one.

\paragraph{The full dynamics.}
Putting everything together, we obtain
\begin{equation}
    \dot s \simeq \frac{\alpha}{n_B} \sum_b \frac{\sqrt{d}}{\lVert h_b \rVert} \Big[ \sin^2\theta_b\, \gamma_{\mathrm{sh},b} - \sin\theta_b \cos\theta_b\, \gamma_{\mathrm{ind},b} \Big],
    \label{eq:app_shift_full}
\end{equation}
where the only approximation is Equation~\ref{eq:meanwrite}.
Without normalization, $\delta_b = W_2^\top \big(p_b - \mathbf{1}_{y_b}\big)$ and $\hat u_{\mathrm{sh}}^\top \delta_b = -\gamma_{\mathrm{sh},b}$, so
\begin{equation}
    \dot s \simeq \alpha\, \bar\gamma_{\mathrm{sh}}, \quad \text{with } \bar\gamma_{\mathrm{sh}} = \frac{1}{n_B}\sum_b \gamma_{\mathrm{sh},b} .
\end{equation}
The term in $\gamma_{\mathrm{ind},b}$ only exists because the normalization removes the radial component of the error, and it is the only term whose sign is controlled by how well the individual parts of the new facts are learned.
Finally, both margins vanish when the new facts are learned: as $p_b \to \mathbf{1}_{y_b}$, $\gamma_{\mathrm{sh},b} \to 0$ and $\gamma_{\mathrm{ind},b} \to 0$.
The force that writes the common shift and the force that withdraws it therefore both fade as the errors of the new facts vanish, with and without normalization.

\subsection{Reduction under the common-coordinate approximation}
\label{app:shift_reduced}

Equation~\ref{eq:app_shift_full} depends on every new fact through $\theta_b$, $\lVert h_b \rVert$, $\gamma_{\mathrm{sh},b}$ and $\gamma_{\mathrm{ind},b}$.
To obtain a closed form, we use Assumption~4: all new states have the same coordinate $s$ along $\hat u_{\mathrm{sh}}$ and the same individual norm $\sigma$.
This is motivated by the fact that the shared part of every hidden state, $\sqrt{\alpha}\, W_1 \mu$, is the same for old and new facts.
Under this assumption,
\begin{equation}
    \lVert h_b \rVert = \sqrt{s^2 + \sigma^2}, \quad \cos\theta_b = \frac{s}{\sqrt{s^2+\sigma^2}}, \quad \sin\theta_b = \frac{\sigma}{\sqrt{s^2+\sigma^2}},
\end{equation}
so that
\begin{equation}
    \frac{\sqrt{d}}{\lVert h_b \rVert}\sin^2\theta_b = \frac{\sqrt{d}\,\sigma^2}{(s^2+\sigma^2)^{3/2}} \quad \text{and} \quad \frac{\sqrt{d}}{\lVert h_b \rVert}\sin\theta_b\cos\theta_b = \frac{\sqrt{d}\,\sigma s}{(s^2+\sigma^2)^{3/2}} .
\end{equation}
Plugging these into Equation~\ref{eq:app_shift_full} and averaging the margins over the new facts yields Equation~\ref{eq:shift},
\begin{equation}
    \dot s = \alpha\,\sqrt{d}\, \frac{\sigma \left( \sigma \bar\gamma_{\mathrm{sh}} - s \bar\gamma_{\mathrm{ind}} \right)}{\left( s^2 + \sigma^2 \right)^{3/2}} .
\end{equation}
This one-dimensional equation has the following properties, which underlie the discussion of the main text:
\begin{itemize}
    \item[--] \textbf{The sign of $\dot s$ is the sign of $\sigma \bar\gamma_{\mathrm{sh}} - s \bar\gamma_{\mathrm{ind}}$.} If $\bar\gamma_{\mathrm{ind}} < 0$ and $\bar\gamma_{\mathrm{sh}} \geq 0$, then $\dot s > 0$: the shift can only grow and there is no fixed point.
    \item[--] \textbf{The shift reinforces itself while it is small.} The weight of the $\bar\gamma_{\mathrm{ind}}$ term, $s\sigma / (s^2+\sigma^2)^{3/2}$, increases with $s$ for $s < \sigma/\sqrt{2}$ and decreases beyond. While the new facts are confidently wrong ($\bar\gamma_{\mathrm{ind}} < 0$) and the shift is small, a larger shift therefore makes the shift grow faster.
    \item[--] \textbf{Once the new facts are right, the shift is pulled back.} If $\bar\gamma_{\mathrm{ind}} > 0$, the equation has a unique fixed point $s^* = \sigma \bar\gamma_{\mathrm{sh}} / \bar\gamma_{\mathrm{ind}}$, which is stable since
    \begin{equation}
        \frac{\partial \dot s}{\partial s}\Big|_{s = s^*} = -\frac{\alpha\,\sqrt{d}\,\sigma\, \bar\gamma_{\mathrm{ind}}}{(s^{*2}+\sigma^2)^{3/2}} < 0 .
    \end{equation}
    As $\bar\gamma_{\mathrm{sh}}$ quickly becomes small, $s^*$ is small and $s$ relaxes towards it.
    \item[--] \textbf{The pull stops with the errors.} The restoring term is proportional to $\bar\gamma_{\mathrm{ind}}$, which vanishes as the new facts are learned. Nothing in the equation brings $s$ back to its initial value: it stops wherever it is when the errors of the new facts vanish.
    \item[--] \textbf{Without normalization, nothing pulls the shift back.} The equation becomes $\dot s = \alpha\,\bar\gamma_{\mathrm{sh}}$, which contains neither $\bar\gamma_{\mathrm{ind}}$ nor $s$.
\end{itemize}

\subsection{Empirical validation}
\label{app:shift}

\begin{figure}[t]
    \centering
    \includegraphics[width=\linewidth]{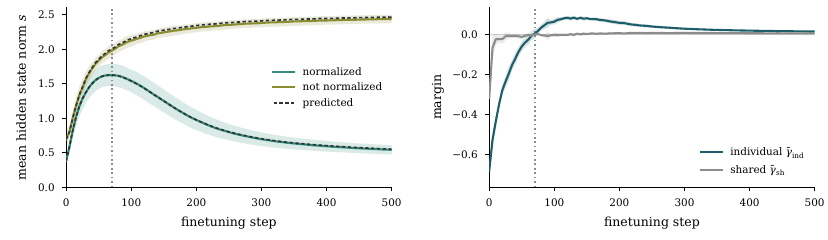}
    \caption{\textbf{Equation~\ref{eq:app_shift_full} reproduces the dynamics of the common shift.} (left) Length $\lVert \bar h_A \rVert$ of the old facts' mean hidden state, measured and obtained by integrating Equation~\ref{eq:app_shift_full} (with normalization) or $\dot s = \alpha\bar\gamma_{\mathrm{sh}}$ (without normalization) from the point at which the mean state has aligned with the shared direction, after which its length follows the same dynamics as its displacement $s$. (right) Average shared margin $\bar\gamma_{\mathrm{sh}}$ and individual margin $\bar\gamma_{\mathrm{ind}}$ of the new facts, with normalization. The dotted line marks the peak; ten seeds.}
    \label{fig:app_shift_validation}
\end{figure}

We verify that Equation~\ref{eq:app_shift_full} describes the actual model, without the common-coordinate approximation.
We follow the finetuning run, trained with minibatches of 32 new facts, and at every step measure $\theta_b$, $\lVert h_b \rVert$, $\gamma_{\mathrm{sh},b}$ and $\gamma_{\mathrm{ind},b}$ on that step's minibatch.
We evaluate the right-hand side of Equation~\ref{eq:app_shift_full} from these measurements and integrate it with an explicit Euler scheme whose time step is the learning rate $\eta$ (Assumption~1), starting from the measured value of $s$ once the mean state has aligned with the shared direction.
The integration includes the second-order term of a finite step, which lengthens the mean state by $\lVert\Delta_\perp\rVert^2/2s$, where $\Delta_\perp$ is the part of the step orthogonal to $\hat u_{\mathrm{sh}}$, and vanishes as $\eta \to 0$.
We proceed in the same way for the model without normalization, with $\dot s = \alpha\,\bar\gamma_{\mathrm{sh}}$.
The prediction matches the measured $s$ to within $1\%$ of its range with normalization and $2\%$ without (Figure~\ref{fig:app_shift_validation}).

\subsection{The mechanism, step by step}
\label{app:toy_checks}

Section~\ref{sec:toy} explains spurious forgetting through a chain of claims about the
minimal model. Here we check each of them in the base configuration, over ten seeds; dotted
lines mark the bottom of the collapse.

\begin{figure}[h]
  \centering
  \includegraphics[width=\linewidth]{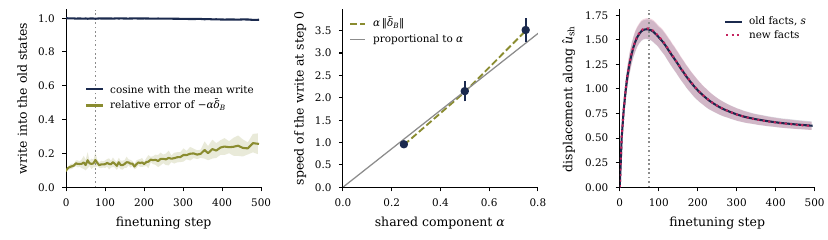}
  \caption{\textbf{How the shift is written.} (left) Cosine between the update of each old
  hidden state and their mean update, and relative error of Equation~\ref{eq:meanwrite}.
  (middle) Speed of the update at the start of finetuning against $\alpha$.
  (right) Displacement of the old and new hidden states along the shared direction.}
  \label{fig:toy_checks_write}
\end{figure}

\noindent\textbf{How the shift is written.}
Equation~\ref{eq:meanwrite} replaces the update of each old hidden state by a single vector,
$-\alpha\bar\delta_B$, shared by all of them. The exact update of every old state indeed
points along the mean update, and $-\alpha\bar\delta_B$ reproduces it to within $16\%$ until
the end of the collapse (Figure~\ref{fig:toy_checks_write}, left); the fact-specific terms it
neglects only grow later, as the erosion sets in. The speed of this write grows almost in
proportion to $\alpha$ (middle), and the new facts move along the shared direction exactly as
far as the old ones (right), so the shift is built and undone on both together.

\begin{figure}[h]
  \centering
  \includegraphics[width=\linewidth]{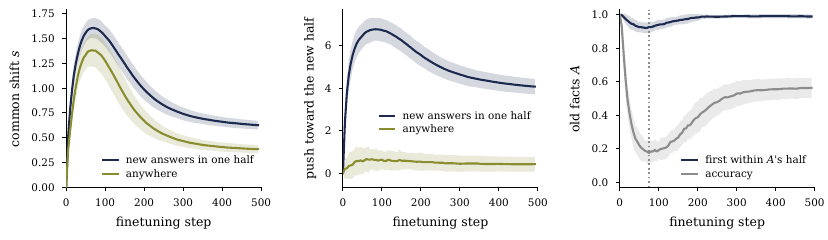}
  \caption{\textbf{The shift favors the half of the new answers.} (left) Common shift with the
  new answers in one half of the vocabulary or anywhere. (middle) How far it pushes the logits
  of $A$ toward the half of the new answers. (right) Old facts whose correct answer is still
  ranked first within their own half, against their accuracy.}
  \label{fig:toy_checks_region}
\end{figure}

\noindent\textbf{Where the shift points.}
When the new answers are drawn from the entire vocabulary, the old states still move
together, nearly as far (Figure~\ref{fig:toy_checks_region}, left), but the shift no longer
favors either half of the vocabulary: its push on the logits of $A$ toward the half of the
new answers is ten times smaller (middle). With the new answers in one half, the facts of $A$
lose to that half but keep their order within their own: at the bottom of the collapse,
$93\%$ of them still rank their correct answer first within the half of $A$, against $17\%$
accuracy (right).

\begin{figure}[h]
  \centering
  \includegraphics[width=\linewidth]{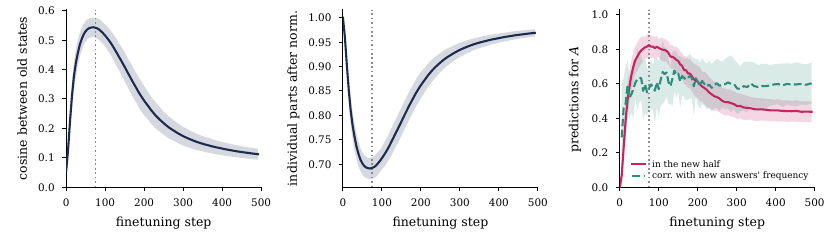}
  \caption{\textbf{Normalization rotates the old states toward the shift.} (left) Mean cosine
  between the old hidden states. (middle) Norm of their individual parts after normalization,
  relative to the start of finetuning. (right) Share of the predictions for $A$ that fall in
  the half of the new answers, and their correlation with how often each answer occurs among
  the new facts.}
  \label{fig:toy_checks_norm}
\end{figure}

\noindent\textbf{What normalization does.}
As the shift grows, normalization rotates the old states toward it: their mean pairwise
cosine rises from $0.06$ to $0.54$ at the bottom of the collapse
(Figure~\ref{fig:toy_checks_norm}, left), and their individual parts lose about a third of
their size (middle). Both recover as the shift is withdrawn. Meanwhile the predictions for $A$
move to the half of the new answers, $82\%$ of them at the bottom of the collapse, favoring
the answers that are most common among the new facts (right).

\begin{figure}[h]
  \centering
  \includegraphics[width=\linewidth]{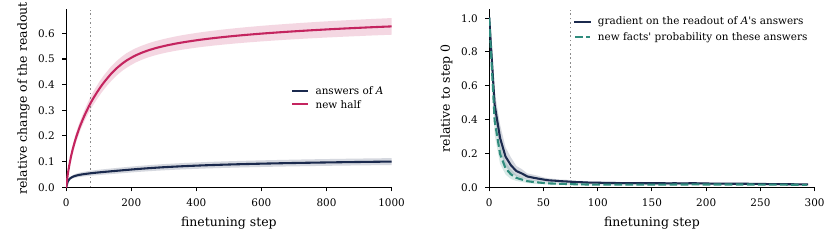}
  \caption{\textbf{The readout of the answers of $A$ is barely touched.} (left) Relative change
  of the readout weights of the answers of $A$ and of the half of the new answers. (right)
  Gradient on the readout weights of the answers of $A$, and the probability the model puts on
  these answers for the new facts, both relative to the start of finetuning.}
  \label{fig:toy_checks_readout}
\end{figure}

\noindent\textbf{The readout of $A$ is barely touched.}
The readout weights of the answers of $A$ change by about $5\%$ during the collapse and $10\%$
by the end of finetuning, against more than $60\%$ for the half of the new answers
(Figure~\ref{fig:toy_checks_readout}, left). They receive gradient only while the model puts
probability on these answers for the new facts, which stops within a few dozen steps (right).

\begin{figure}[h]
  \centering
  \includegraphics[width=\linewidth]{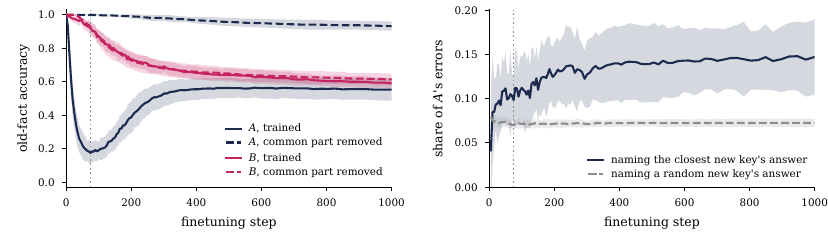}
  \caption{\textbf{The shift hides $A$; the drift erodes $A$ and $B$.} (left) Accuracy of $A$ and
  $B$ as trained and with the common part of the change of their logits removed. (right) Among
  the facts of $A$ recalled wrongly, the share that names the answer of the closest new key,
  against that of a random new key.}
  \label{fig:toy_checks_survive}
\end{figure}

\noindent\textbf{What survives and what erodes.}
Removing the common part of the change of the logits restores $A$ almost entirely at the
bottom of the collapse, and slightly less as finetuning proceeds; for $B$, whose answers are
those of the new facts, it changes almost nothing, and $B$ erodes steadily after the collapse
(Figure~\ref{fig:toy_checks_survive}, left). The fact-specific drift behind this erosion is not
neutral: when a fact of $A$ is recalled wrongly, it names the answer of the new fact whose key
is closest to its own twice as often as that of a random new fact (right).

\subsection{The old facts survive the common shift}
\label{app:shared_row}

\begin{figure}[t]
    \centering
    \includegraphics[width=\linewidth]{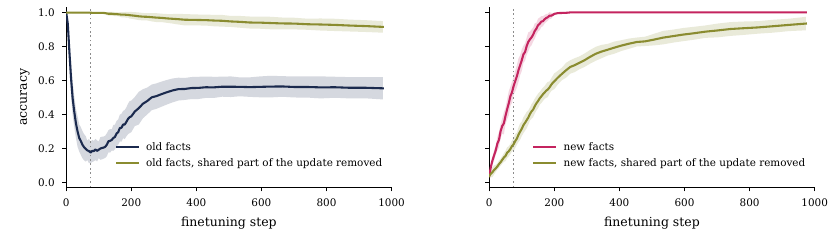}
    \caption{\textbf{Removing the shared part of the update restores the old facts.} Accuracy on the old (left) and new (right) facts as trained and with $\Delta W_1\,\mu\mu^\top$ removed from the store, the readout kept as trained. The dotted line marks the trough; ten seeds.}
    \label{fig:app_shared_row}
\end{figure}

The common shift of the old states is written by the part of the update that acts on the shared key direction, $\Delta W_1\,\mu\mu^\top$ (Equation~\ref{eq:split}).
At every checkpoint, we remove this part from the store, $W_1(t) \leftarrow W_1(0) + \Delta W_\perp(t)$ with $\Delta W_\perp = \Delta W_1 (I - \mu\mu^\top)$, keep the readout as trained, and evaluate the edited model; training itself is not modified.
At the trough, the edited model recalls the old facts with accuracy $1.00$, against $0.17$ for the trained model, and after about 1{,}000 steps with accuracy $0.91$, against $0.55$ (Figure~\ref{fig:app_shared_row}).
The remaining decline is the erosion carried by $\Delta W_\perp$ (Section~\ref{app:laws}).
The edit also slows the new facts, which reach $0.59$ at step 200 instead of $0.99$, since the shared part of the update also carries what the new facts have in common.

\begin{figure}[t]
    \centering
    \includegraphics[width=\linewidth]{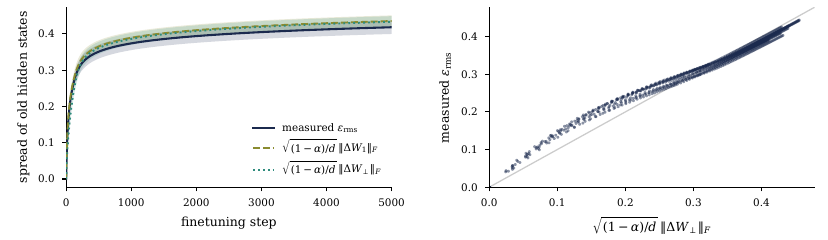}
    \caption{\textbf{The spread of the old facts is set by the Frobenius norm of the update.} (left) Measured spread $\varepsilon_\mathrm{rms}$ of the old hidden states during finetuning, compared with $\sqrt{(1-\alpha)/d}\,\lVert \Delta W_1 \rVert_F$ and with $\sqrt{(1-\alpha)/d}\,\lVert \Delta W_\perp \rVert_F$ (Equation~\ref{eq:erosion}). (right) Measured against predicted; the prediction runs slightly low only while the shared column of the update is large. Ten seeds.}
    \label{fig:app_erosion}
\end{figure}

\subsection{The analyses of the Transformer and the language model, in the minimal model}
\label{app:toy_mirror}

Sections~\ref{sec:transformer} and~\ref{sec:llm} test the mechanism with two analyses that
need no access to a model's internals: splitting the displacement of the old facts and the
change of their logits into a common and a fact-specific part, and removing the top direction
of each weight update. We run both on the minimal model, where the mechanism is known, to check
that they detect it.

\begin{figure}[h]
  \centering
  \includegraphics[width=\linewidth]{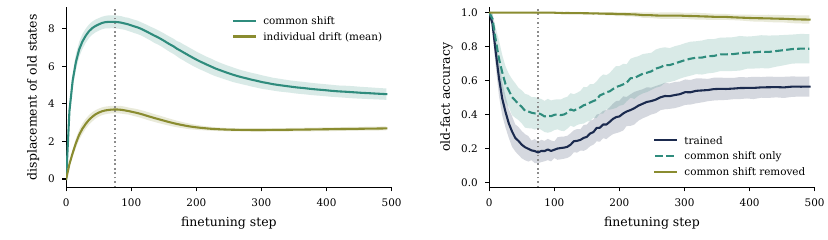}
  \caption{\textbf{In the minimal model, removing the common shift removes the collapse.} (left) Common shift and fact-specific drift of the old hidden states after normalization. (right) Old-fact accuracy as trained, with only the common part of the change of their logits, and with it removed; the dotted line marks the bottom of the collapse.}
  \label{fig:toy_mirror}
\end{figure}

\noindent\textbf{The split in the minimal model.}
We measure the old hidden states after the normalization, at the input of the readout, as in
the Transformer. Their common shift grows fast, peaks at the bottom of the collapse and then
partly retracts (Figure~\ref{fig:toy_mirror}, left). The fact-specific drift stays smaller for
much longer than in the Transformer, and overtakes the shift only after about $10^4$ steps:
once the new facts are learned, both the withdrawal of the shift and the growth of the drift are
driven by their vanishing errors, so the remaining gap closes only logarithmically
(Appendix~\ref{app:laws}). Subtracting the common part of the change of the logits restores the
old facts at the bottom of the collapse, from $0.17$ to $1.00$, while keeping only that part
produces most of the collapse (Figure~\ref{fig:toy_mirror}, right), as in the Transformer.

\begin{figure}[h]
  \centering
  \includegraphics[width=\linewidth]{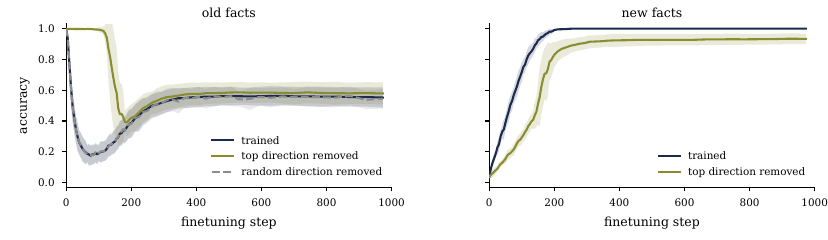}
  \caption{\textbf{Removing the top direction of the update in the minimal model.} Accuracy on
  the old (left) and new (right) facts as trained, with the top singular component of the
  update of $W_1$ removed, and, for the old facts, with a random rank-one matrix of the same
  norm removed. }
  \label{fig:toy_svd}
\end{figure}

\noindent\textbf{Removing the top direction of the update.}
As for the language model, we subtract from the store $W_1$ the top singular component of its
update $W_1(t)-W_1(0)$, or, as a control, a random rank-one matrix of the same norm, and keep
the readout as trained. At the bottom of the collapse, the edit restores the old facts entirely,
whereas the control leaves them where training did (Figure~\ref{fig:toy_svd}, left). The top
direction is then the shared one: its input direction is aligned with $\mu$ (cosine $0.99$),
so the edit removes the part of the update that writes the common shift
(Appendix~\ref{app:shared_row}). It also slows the learning of the new facts (right), which
use the same direction. Once the shift recedes, the top direction no longer carries it
(cosine $0.11$ by step 200), and the edit no longer helps. Compared at the same accuracy on
the new facts, however, the edited model always recalls more old facts than the trained one,
so the edit does not simply rewind finetuning, as in the language model.

\subsection{Fact-specific displacements}
\label{app:laws}

\paragraph{Size of the fact-specific displacements.}
Recall from Equation~\ref{eq:app_eps} that $\varepsilon_a = \sqrt{1-\alpha}\,\Delta W_1 (e_a - \bar e_A)$.
Under Assumption~3, $\Delta W_1$ is fixed with respect to the old keys, and we can take the expectation over them.
For a fixed matrix $M$ and a vector $u$ with $\mathbb{E}[u u^\top] = \Sigma$, $\mathbb{E}\lVert M u \rVert^2 = \mathrm{tr}(M \Sigma M^\top)$.
The $e_a$ are independent and uniform on the unit sphere, so $\mathbb{E}[e_a e_a^\top] = I/d$ and
\begin{equation}
    \mathbb{E}\left[ (e_a - \bar e_A)(e_a - \bar e_A)^\top \right] = \left[ \left(1 - \frac{1}{n_A}\right)^2 + \frac{n_A - 1}{n_A^2} \right] \frac{I}{d} = \left( 1 - \frac{1}{n_A} \right) \frac{I}{d} ,
\end{equation}
from which
\begin{equation}
    \mathbb{E}\left[ \varepsilon_\mathrm{rms}^2 \right] = \frac{1-\alpha}{d} \left( 1 - \frac{1}{n_A} \right) \lVert \Delta W_1 \rVert_F^2 .
\end{equation}
Each old fact reads the update through its own random key, and a random direction captures, on average, a fraction $1/d$ of its squared Frobenius norm.

\paragraph{The shared and non-shared parts of the update.}
We split the update into its column along $\mu$ and the rest, $\Delta W_1 = \Delta W_\perp + (\Delta W_1 \mu)\, \mu^\top$ with $\Delta W_\perp := \Delta W_1 (I - \mu\mu^\top)$.
The two parts are orthogonal for the Frobenius inner product, so $\lVert \Delta W_1 \rVert_F^2 = \lVert \Delta W_\perp \rVert_F^2 + \lVert \Delta W_1 \mu \rVert^2$, and, neglecting the $1/n_A$ correction,
\begin{equation}
    \varepsilon_\mathrm{rms}^2 \simeq \frac{1-\alpha}{d} \left( \lVert \Delta W_\perp \rVert_F^2 + \lVert \Delta W_1 \mu \rVert^2 \right).
\end{equation}
The second term is the shared column of the update, read through the small random component $\mu^\top e_a$ of each old key.
It only matters while the shared write makes up a large part of the update, that is, around the peak of the common shift.
Dropping it gives
\begin{equation}
    \varepsilon_\mathrm{rms} \simeq \sqrt{\frac{1-\alpha}{d}}\, \lVert \Delta W_\perp \rVert_F .
    \label{eq:erosion}
\end{equation}
Figure~\ref{fig:app_erosion} compares the measured spread with both predictions.

\paragraph{Why the fact-specific displacements are not undone.}
Subtracting the mean write from Equation~\ref{eq:app_ha}, the fact-specific displacements evolve as
\begin{equation}
    \dot \varepsilon_a = -\frac{1}{n_B} \sum_b k_b^\top (k_a - \bar k_A)\, \delta_b = -\frac{\sqrt{1-\alpha}}{n_B} \sum_b k_b^\top (e_a - \bar e_A)\, \delta_b .
\end{equation}
The coefficients are random, of order $d^{-1/2}$, and independent of the old fact's own state and answer.
Unlike Equation~\ref{eq:app_shift_full}, this expression contains no term that depends on $\varepsilon_a$ itself and could pull it back.
The contribution of the normalization to $\delta_b$ is present as well, but it enters with the same random coefficients.
The fact-specific displacements therefore keep accumulating as long as the new facts produce errors.

\subsection{Why recovery is incomplete}
\label{app:readout}

Two effects keep the old facts from recovering fully.
The first follows from Section~\ref{app:shift_full}: the force that withdraws the common shift is proportional to the margins of the new facts and vanishes with their errors.
The second comes from the readout, which we have ignored so far.

\begin{figure}[t]
    \centering
    \includegraphics[width=\linewidth]{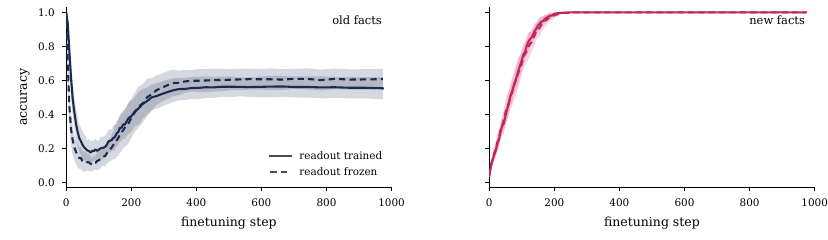}
    \caption{\textbf{Part of the bias towards the new answer region is stored in the readout.} Accuracy on the old and new facts during finetuning, with the readout trained and frozen; the learning rate of the frozen run is matched so that the new facts are learned at the same pace. Ten seeds.}
    \label{fig:app_readout}
\end{figure}

Under the gradient flow dynamics, Equation~\ref{eq:app_gradW2} implies that, for a fixed normalized old state $x_a$, the logits change as
\begin{equation}
    \dot z_a = \dot W_2 x_a = \frac{1}{n_B} \sum_b \big(x_b^\top x_a\big) \big(\mathbf{1}_{y_b} - p_b\big).
\end{equation}
This has the same structure as Equation~\ref{eq:app_ha}, with the overlap between normalized states in place of the overlap between keys.
The normalized states share the direction of the common shift, so the overlaps $x_b^\top x_a$ have a common positive part.
The readout therefore raises the logits of the new answers for every old fact at once, and lowers those of the answers the model currently predicts for the new facts.
Unlike the write into $W_1$, this update has no normalization term: nothing in it depends on the margins of the new facts or on what the readout has already written.
It stops when $p_b \to \mathbf{1}_{y_b}$ but never reverses, so the bias towards the new answer region that it stores remains after the new facts are learned.
Freezing the readout during finetuning removes this channel and raises the recovered accuracy of the old facts from $0.56$ to $0.61$ (Figure~\ref{fig:app_readout}).

\begin{figure}[t]
    \centering
    \includegraphics[width=\linewidth]{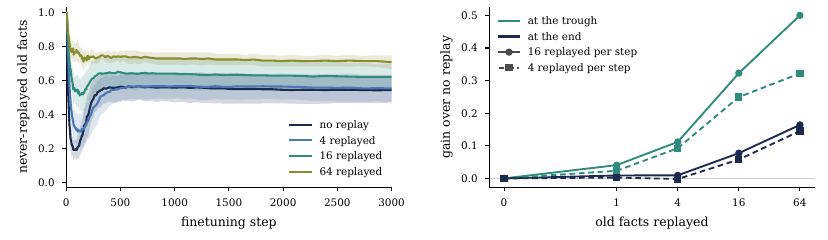}
    \caption{\textbf{Replaying a few old facts protects the others from the collapse, but not from erosion.} (left) Accuracy of the old facts that are never replayed, when 0, 4, 16 or 64 of the 128 old facts are replayed in every batch (16 replayed examples per step; five seeds). (right) Gain over no replay at the trough and at the end of finetuning (3{,}000 steps), with 16 or 4 replayed examples per step.}
    \label{fig:app_replay}
\end{figure}

\subsection{Interventions in the minimal model}
\label{app:interventions}

\paragraph{Projecting out the shared direction.}
\label{app:projection}
Removing the component of the update along the shared key direction at every training step, $\Delta W_1 \leftarrow \Delta W_1 (I - \mu\mu^\top)$, prevents the collapse entirely, and the new facts are still learned.
The old facts nevertheless end no better off than without the projection (accuracy $0.45$ against $0.51$ after 5{,}000 steps, ten seeds), because erosion lies in the fact-specific part of the update, which the projection leaves untouched.

\paragraph{Replaying a few old facts.}
\label{app:replay}
Rehearsing a subset of old associations can restore the others \citep{hinton1987using}.
We replay a fixed random subset of the old facts in every finetuning batch of 32 new facts and measure the old facts that are never replayed (Figure~\ref{fig:app_replay}).
Replaying 16 of the 128 old facts lifts their trough from $0.17$ to $0.50$, as the replayed facts push back on the shift that all old facts share, but raises their final accuracy only from $0.54$ to $0.62$: erosion is specific to each fact and is barely affected by replaying others.
A single replayed fact has almost no effect, as it is soon answered correctly through its own key and stops pushing back on the shared direction.

\clearpage
\newpage
\section{The controlled Transformer}
\label{app:transformer}

\subsection{The old facts keep their ranking within their region}
\label{app:rank}

\begin{figure}[h]
  \centering
  \includegraphics{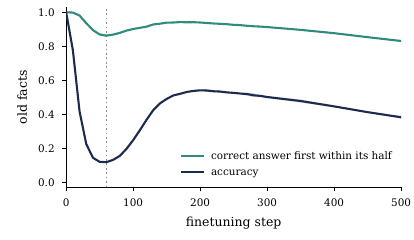}
  \caption{\textbf{The collapse leaves the old facts' ranking within their region intact.} Accuracy of the old facts of $A$ over all answers, and fraction of them whose correct answer still ranks first among the answers of its own half, for the transformer of Figure~\ref{fig:one} (right; three seeds). The dotted line marks the trough.}
  \label{fig:app_rank}
\end{figure}

For each old fact of $A$, we rank its correct value among the values of its own half of the answers, and report the fraction of old facts for which it ranks first, alongside their accuracy over all answers (Figure~\ref{fig:app_rank}).
At the bottom of the collapse, only $12\%$ of the old facts are recalled, yet $86\%$ still rank their correct answer first within their half: the old facts lose to the answers of the other half, not to one another.
The fraction then declines only slowly, to $83\%$ after 500 steps, as the fact-specific drift erodes the old facts.

\subsection{Timing across learning rates and batch sizes}
\label{app:dose}

We finetune the transformer of Figure~\ref{fig:one} under five conditions that
change how fast it learns: the reference learning rate and batch size of 256, half
and twice the learning rate, and batches of 128 and 512 at the reference learning
rate. Each condition is run with three seeds, and both old and new facts are evaluated every 2 steps on held-out phrasings. We define the
trough as the point of lowest old-fact accuracy before the recovery.

\begin{figure}[h]
  \centering
  \includegraphics[width=\linewidth]{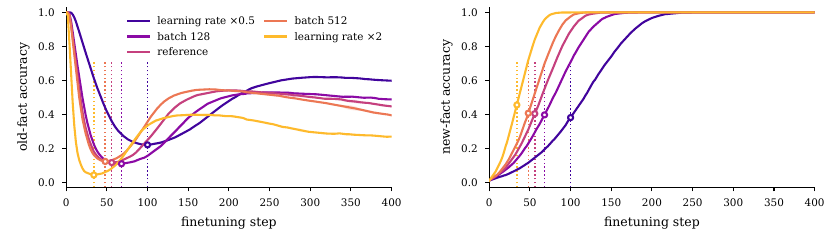}
  \caption{\textbf{The end of the collapse follows the learning of the new facts.} Old-fact
  (left) and new-fact (right) accuracy under five finetuning conditions; circles and dotted
  lines mark each condition's trough. The troughs span steps 34 to 100, yet the new facts are
  at 36--44\% accuracy at every one of them. Three seeds, evaluated every 2 steps.}
  \label{fig:dose}
\end{figure}

Across the five conditions, the trough moves from step 34 to step 100, a factor of about three in training steps. At every trough, however, the new facts have reached nearly
the same stage of learning, between $36\%$ and $44\%$ accuracy (Figure~\ref{fig:dose}). The end of the collapse therefore follows the learning of the new facts rather than a fixed amount of optimization, as the minimal model predicts. This holds for a fixed set of new facts; with fewer of them, the trough falls later in their learning (Appendix~\ref{app:setsize}).

\subsection{Timing across the number of new facts}
\label{app:setsize}

Varying the learning rate or the batch size changes how fast the new facts are learned, but not how many there are. To vary the latter, we finetune on only the first 125 or 250 of the 500 individuals of $B'$, keeping everything else as in the reference condition of Appendix~\ref{app:dose}, and evaluate every 2 steps.

\begin{figure}[t]
  \centering
  \includegraphics[width=\linewidth]{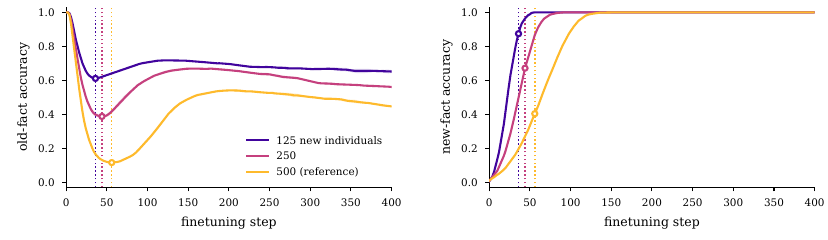}
  \caption{\textbf{With fewer new facts, the collapse ends later in their learning.}
  Old-fact (left) and new-fact (right) accuracy of the Transformer finetuned on 125, 250 or
  500 new individuals; circles and dotted lines mark the bottom of each collapse. Three
  seeds.}
  \label{fig:setsize}
\end{figure}

The bottom of the collapse then no longer falls at the same accuracy on the new facts (Figure~\ref{fig:setsize}). The fewer the new facts, the further they are learned before the old facts start to return: $89\%$ of them are recalled at the bottom with 125 individuals, against $38\%$ with 500. The collapse also becomes shallower, and in the Transformer with MLP blocks (Appendix~\ref{app:mlp}) it nearly disappears, with old-fact accuracy staying above $0.9$ for 125 or 250 individuals.

\begin{figure}[t]
  \centering
  \includegraphics[width=\linewidth]{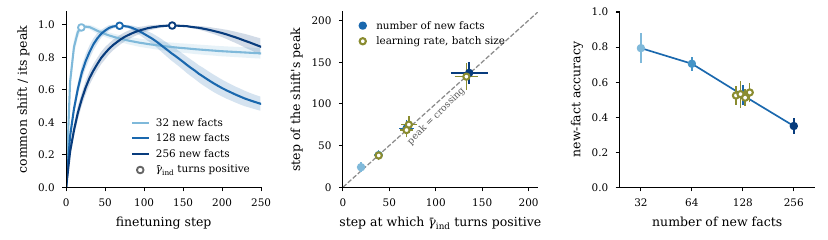}
  \caption{\textbf{In the minimal model, the shift peaks when the individual margin turns
  positive.} (left) Common shift over finetuning with 32, 128 and 256 new facts, scaled to its
  peak; circles mark where $\bar\gamma_{\mathrm{ind}}$ turns positive. (middle) The step of
  the peak against that step, for different numbers of new facts (filled) and for different
  learning rates and batch sizes (open). (right) Accuracy on the new facts at that step:
  it depends on their number, but not on the learning rate or the batch size. Ten seeds.}
  \label{fig:toy_timing}
\end{figure}

The minimal model shows why (Figure~\ref{fig:toy_timing}). Whatever we vary, the learning rate, the batch size or the number of new facts, the common shift peaks when the individual margin $\bar\gamma_{\mathrm{ind}}$ of the new facts turns positive, as Equation~\ref{eq:shift} predicts. How well the new facts are recalled at that moment depends on their number but not on the learning rate or the batch size. A fixed accuracy on the new facts is therefore only a stand-in for the individual margin, and it holds only as long as the number of new facts is fixed.

\begin{table}[t]
\centering\small
\begin{tabular}{lc}
\toprule
Weights restored & Share of the common shift \\
\midrule
Attention value and output & $84\%$ \\
Attention query and key & $11\%$ \\
Input embedding & $2\%$ \\
LayerNorms & $1\%$ \\
Unembedding (control) & $0\%$ \\
\bottomrule
\end{tabular}
\caption{Share of the common shift removed by restoring each group of weights to its
value before finetuning, at the trough (step 50; three seeds, standard deviation below 0.3 percentage points).}
\label{tab:localization}
\end{table}
\subsection{Localizing the common shift}
\label{app:localization}

To find which weights carry the common shift, we take the checkpoint at the end of the collapse
(step 50), restore one group of weights to its value before finetuning, and recompute
the final hidden states of the old facts. We measure the common shift $\Delta \bar h$ of
Section~\ref{sec:transformer} with and without the restoration, and report the norm
of the difference as a fraction of $\lVert \Delta \bar h \rVert$, averaged over the six
attributes. The groups partition all parameters, so their shares can be compared directly; each is measured with only that group restored, so they need not add up to $100\%$. Nothing is retrained, so the learning of the new facts is
unaffected by construction.

The attention value and output weights carry $84\%$ of the shift, and the query and
key weights $11\%$ (Table~\ref{tab:localization}); after the recovery (step 200),
the value and output weights still carry $85\%$. The unembedding is applied after the point at which the shift is
measured and serves as a control: restoring it leaves the shift unchanged, as it
must.

\subsection{Transformer with MLP blocks}
\label{app:mlp}

We repeat every analysis of Section~\ref{sec:transformer} with a Transformer that has MLP blocks of width $4d$ and is otherwise identical: same data, same size, same pretraining and finetuning. The picture is the same, with one difference that matters: the fact-specific drift takes a larger part in the collapse.

\begin{figure}[t]
  \centering
  \includegraphics[width=\linewidth]{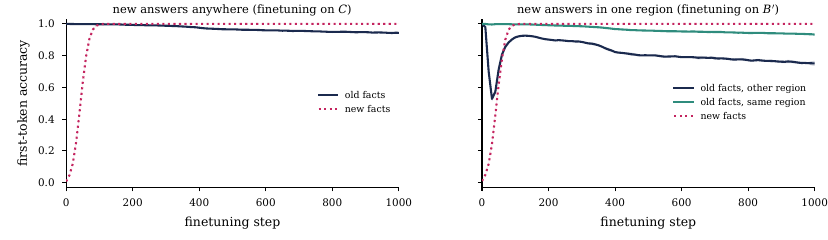}
  \caption{\textbf{The Transformer with MLP blocks reproduces Figure~\ref{fig:one}.}
  Accuracy of the old and new facts when finetuning on $C$ (left) and on $B'$ (right).}
  \label{fig:mlp_one}
\end{figure}

\noindent\textbf{Collapse, recovery and erosion.}
Finetuning on $C$ lowers recall of the old facts only slowly, to $0.94$ after 1{,}000 steps (Figure~\ref{fig:mlp_one}, left). Finetuning on $B'$ makes the old facts of $A$ collapse to $0.53$ at step 30, recover to $0.93$ and then erode to $0.75$, while those of $B$, whose answers the new facts share, barely move during the collapse (right).
\begin{figure}[t]
  \centering
  \includegraphics[width=\linewidth]{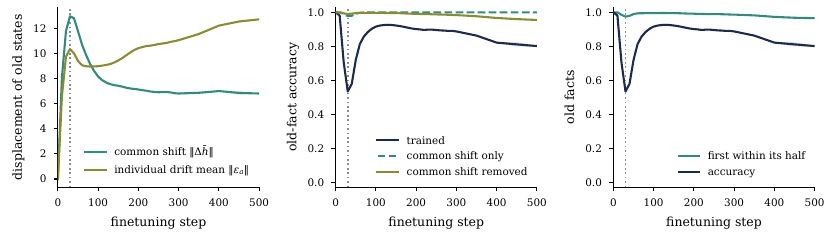}
  \caption{\textbf{With MLP blocks, removing the common shift still removes the collapse.}
  (left) Common shift and fact-specific drift of the old hidden states. (middle) Old-fact
  accuracy as trained, with only the common part of the change of their logits, and with it
  removed. (right) Old facts whose correct answer still ranks first within their own half,
  against their accuracy. Three seeds; the dotted line marks the bottom of the collapse.}
  \label{fig:mlp_split}
\end{figure}

\noindent\textbf{Common shift and fact-specific drift.}
As in Figure~\ref{fig:transformer}, the common shift grows fast, peaks at the bottom of the collapse and then partly retracts, while the fact-specific drift keeps growing and overtakes it after about 90 steps (Figure~\ref{fig:mlp_split}, left). Subtracting the common part of the change of the logits removes the collapse, raising old-fact accuracy at the bottom from $0.53$ to $0.99$ (middle), and the old facts keep their order within their own half: $97\%$ of them still rank their correct answer first within it (right).

\begin{figure}[h]
  \centering
  \includegraphics[width=\linewidth]{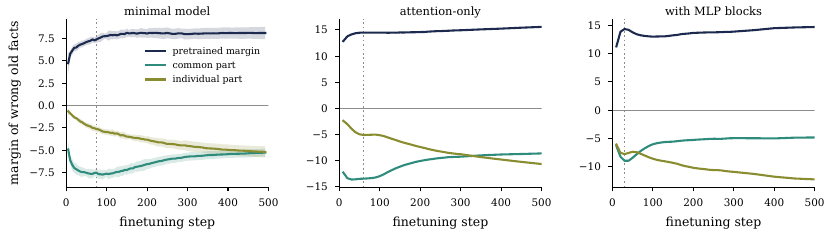}
  \caption{\textbf{What turns an old fact wrong.} For the old facts recalled wrongly, the
  margin of the correct answer over its strongest competitor, split into its pretrained value
  and the contributions of the common and fact-specific parts of the change of the logits.}
  \label{fig:margin_split}
\end{figure}

\noindent\textbf{The common shift turns the drift into a collapse.}
Keeping only the common part lowers recall less here than in the attention-only Transformer
(Figure~\ref{fig:mlp_split}, middle), while removing it still restores the old facts. Both
models therefore collapse in the way Section~\ref{sec:transformer} describes: the common shift
brings the old facts close to losing to the new answer region, and the fact-specific drift
pushes them across. The two differ only in how the work is shared. Splitting the margin of each
old fact recalled wrongly, its correct logit minus that of its strongest competitor, into its
pretrained value and the contributions of the common and fact-specific parts
(Figure~\ref{fig:margin_split}), the common part alone almost cancels the pretrained margin in
the minimal model and in the attention-only Transformer, but leaves a larger part of it with
MLP blocks, where the fact-specific drift is larger at the bottom of the collapse. Without the
shift, the drift alone collapses neither model.

\begin{figure}[h]
  \centering
  \includegraphics[width=\linewidth]{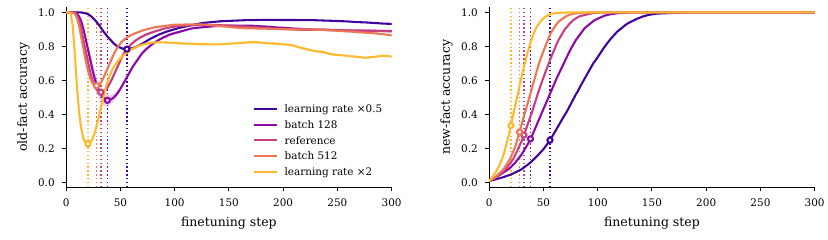}
  \caption{\textbf{Timing in the Transformer with MLP blocks.} Old-fact (left) and new-fact
  (right) accuracy under five learning rates and batch sizes; circles and dotted lines mark
  the bottom of each collapse. Three seeds evaluated at every 2 steps}
  \label{fig:mlp_timing}
\end{figure}
\noindent\textbf{Where the shift is written.}
Restoring groups of weights as in Appendix~\ref{app:localization}, the MLP weights and the
attention value and output weights each carry about half of the common shift at the bottom of the collapse ($46\%$ and $47\%$), the query and key weights $3\%$, and the embedding and LayerNorms less than $2\%$ each. Both MLP and attention value and output weights write content into the residual stream, so the shift is again written into the associative memories of the network rather than into where attention goes. Adding MLP blocks does not move
the shift to a new place, but spreads it over a second memory: the update that all new facts share is written wherever the network stores associations, in proportion to how much each store takes part in learning them. The query and key weights, which decide where attention goes, contribute little during the collapse and somewhat more after the recovery, as in the attention-only Transformer, where their share also grows once the fact-specific drift dominates. The collapse is thus a change in what the network writes about each fact, not in which tokens it reads.

\noindent\textbf{Timing.}
Changing the learning rate and the batch size as in Appendix~\ref{app:dose} moves the bottom of the collapse from step 20 to step 56, while the new facts are recalled with $23$ to $33\%$ accuracy at every bottom (Figure~\ref{fig:mlp_timing}). As in the attention-only Transformer, the end of the collapse follows the learning of the new facts rather than the number of steps: a faster optimizer reaches the bottom sooner, but at almost the same point in learning the new facts. This point is not shared across architectures, however: with MLP blocks, the
collapse ends when fewer new facts are recalled than without them. What the two models have in common is therefore not a threshold on new-fact accuracy, but the mechanism that sets it, the individual margin of the new facts turning positive, whose relation to accuracy depends on the model and, as Appendix~\ref{app:setsize} shows, on the number of new facts. The depth of the collapse, in contrast, depends on the learning rate, more strongly than without MLP blocks: halving it leaves only a shallow collapse, consistent with Section~\ref{sec:transformer},
where the depth, unlike the timing, lies outside our gradient-flow analysis.

\newpage
\section{The pretrained language model}
\label{app:llm_results}

\subsection{Why recall of the old facts first rises.}
\label{app:llm_rise}
In the first 25 to 30 steps of finetuning, recall of the old facts rises from $0.87$ to about
$0.93$ with both kinds of new facts, although none of them is replayed. Before finetuning, the
model answers 54 of the 422 old facts wrongly, but for 47 of them it already ranks the correct
answer first among the possible answers of the relation, and loses only to a continuation that is
not an answer. Every old fact that becomes correct during the rise is one of these 47 (26 with
synthetic individuals, 32 with real entities), and the rank of the correct answer among the
possible answers does not change. In every finetuning document, the answer immediately follows
the relation, so finetuning raises answers as a whole above other continuations without changing
which answer the model prefers. The collapse sets in once the shift toward the new answers
dominates.

\subsection{Removing one direction of the update}
\label{app:lowrank}

\paragraph{The edit.}
At each evaluation step of a finetuning run, we compute the update
$\Delta W = W(t) - W(0)$ of every weight matrix of the transformer blocks of
OLMo~2~1B: the query, key, value and output matrices of attention and the three
matrices of the MLP, in each of the 16 layers, 112 matrices in total. The embedding
and the unembedding are left untouched. We remove the top singular direction of each
update, $W(t) \leftarrow W(t) - \sigma_1 u_1 v_1^\top$, evaluate the edited model, and
then restore the original weights before training continues, so that the finetuning
run itself is never modified. The removed direction is small: it holds about $15\%$
of the squared norm of a typical update. As a control, we instead remove a random
rank-one matrix of the same norm from each update.

\paragraph{Effect on the old facts.}
Removing the top direction keeps the old facts close to their accuracy before
finetuning through the collapse, for both kinds of new facts, while the random control
leaves them exactly where training left them (Table~\ref{tab:lowrank}). The edit also
keeps the loss on a fixed sample of ordinary text (64 sequences of 512 tokens) close
to its value before finetuning, whereas finetuning alone raises it.

\paragraph{The edit is not a return to an earlier point of training.}
The edit lowers the accuracy on the new facts, which raises the question of whether it
simply moves the model back along its training trajectory. To test this, we compare
the edited model with the trained model at the same accuracy on the new facts: for
each edited checkpoint, we find the step at which training first reached the same
new-fact accuracy, interpolating between evaluations, and read the old-fact accuracy
there. For real entities, at new-fact accuracies of $21\%$, $34\%$ and $66\%$, the
edited model recalls $88\%$, $88\%$ and $79\%$ of the old facts, against $81\%$,
$73\%$ and $72\%$ for the trained model: no point of the training run combines the
edited model's old-fact and new-fact accuracies. The reduction in new-fact accuracy indicates
that the removed direction also carries part of what the new facts share.

\begin{table}[h]
\centering\small
\begin{tabular}{llccccc}
\toprule
 & & \multicolumn{3}{c}{Old facts} & \multicolumn{2}{c}{New facts} \\
\cmidrule(lr){3-5}\cmidrule(lr){6-7}
New facts & Step & trained & top removed & random removed & trained & top removed \\
\midrule
Synthetic individuals & 80  & 0.28 & 0.77 & 0.28 & 0.03 & 0.01 \\
                      & 400 & 0.38 & 0.70 & 0.38 & 0.90 & 0.40 \\
Real entities         & 80  & 0.73 & 0.88 & 0.73 & 0.46 & 0.25 \\
                      & 400 & 0.55 & 0.79 & 0.55 & 1.00 & 0.66 \\
\bottomrule
\end{tabular}
\caption{Accuracy of OLMo~2~1B at the trough (step 80) and after 400 steps of
finetuning, with the top or a random direction of each weight update removed. Before
finetuning, the old facts are recalled with accuracy $0.87$. Small finetuning set for
the synthetic individuals; three seeds.}
\label{tab:lowrank}
\end{table}

\subsection{The same edit in the controlled Transformers}
\label{app:tf_lowrank}

We apply the edit of Section~\ref{sec:llm} to the two Transformers of
Section~\ref{sec:transformer}: at each checkpoint, we remove the top singular component of the
update of every weight matrix in the Transformer blocks (attention and, where present, MLP),
leave the embedding and the unembedding untouched, and evaluate the edited model; as a control,
we remove a random rank-one matrix of the same norm instead. The removed direction holds
$8\%$ of the squared norm of a typical update in the attention-only Transformer and $14\%$ with
MLP blocks, close to the $15\%$ of the language model.

\begin{figure}[h]
  \centering
  \includegraphics[width=\linewidth]{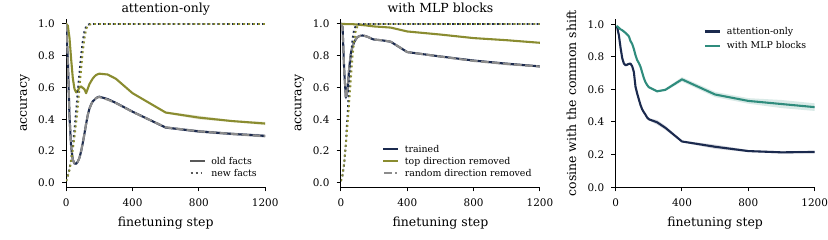}
  \caption{\textbf{Removing the top direction of the update in the controlled Transformers.}
  (left, middle) Accuracy on the old (solid) and new (dotted) facts as trained, with the top
  singular component of every block-weight update removed, and with a random rank-one matrix of
  the same norm removed. (right) Cosine between the change the edit makes to the readout states
  of the old facts and their common shift. Three seeds.}
  \label{fig:tf_lowrank}
\end{figure}

\noindent\textbf{Effect on the old facts.}
At the bottom of the collapse, the edit raises old-fact accuracy from $0.12$ to $0.57$ in the
attention-only Transformer and from $0.54$ to $1.00$ with MLP blocks, whereas the control leaves
it exactly where training did (Figure~\ref{fig:tf_lowrank}, left and middle). The edited model
recalls more old facts than the trained one throughout finetuning, still $0.37$ against $0.29$
and $0.88$ against $0.73$ after 1{,}200 steps. As in the language model, the edit also slows
the learning of the new facts in the attention-only Transformer, as the removed direction
carries part of what they share; with MLP blocks it barely affects them.

\noindent\textbf{Why the top direction carries the shift.}
The common shift is what all new facts write coherently into the old hidden states, so while
it is being written it should dominate the update. We measure how the edit changes the
readout states of the old facts and compare this change with the common shift
(Figure~\ref{fig:tf_lowrank}, right). At the bottom of the collapse the two are aligned,
with cosine $0.75$ in the attention-only Transformer and $0.97$ with MLP blocks. As the
fact-specific drift grows, the top direction aligns less and less with the shift, down to
$0.22$ and $0.49$ after 1{,}200 steps. In the attention-only Transformer, the edit removes only
about a quarter of the shift at the bottom of the collapse and yet restores half of the old
facts: as Appendix~\ref{app:mlp} shows, the common part alone nearly cancels their pretrained
margin in this model, so removing part of it is enough to bring many of them back.

\noindent\textbf{The edit is not a return to an earlier point of training.}
As in Appendix~\ref{app:lowrank}, we compare the edited model with the trained model at the
same accuracy on the new facts. While the new facts are being learned, the edited model recalls
more old facts at every new-fact accuracy: at step 100, $0.60$ against $0.21$ in the
attention-only Transformer and $1.00$ against $0.90$ with MLP blocks. Once all new facts are
recalled, the comparison loses its meaning, as every later checkpoint is matched to the first
step at which training reached full accuracy on the new facts.

\end{document}